\documentclass[onecolumn]{article} 
\usepackage[top=3.6cm, bottom=3.2cm, left=2.5cm, right=2.5cm]{geometry}
\usepackage[colorlinks=true,linkcolor=red,citecolor=blue]{hyperref}
\usepackage{amsthm,amsmath,amssymb}
\usepackage{mathrsfs}
\usepackage{upgreek}

\usepackage{tabularx}
\usepackage{graphicx}
\usepackage{subfigure}

\usepackage{cite}
\usepackage{url}

\begin{document}
\renewcommand{\thefootnote}{\fnsymbol{footnote}}
\title{Enabling Country-Scale Land Cover Mapping with Meter-Resolution Satellite Imagery\footnote{A website is available at \url{https://x-ytong.github.io/project/Five-Billion-Pixels.html}.}}

\author{Xin-Yi Tong$^1$, Gui-Song Xia$^{2, 3}$, Xiao Xiang Zhu$^{4, }$\footnote{Corresponding author.}
\vspace{3mm}
\\
$^1${\em \small Remote Sensing Technology Institute, German Aerospace Center, Germany}\\
$^2${\em \small State Key Laboratory of Information Engineering in Surveying, Mapping and Remote Sensing,}\\ {\em \small Wuhan University, China}\\
$^3${\em \small National Engineering Research Center for Multi-media Software, School of Computer Science}\\ {\em \small and Institute of Artificial Intelligence, Wuhan University, China}\\
$^4${\em \small Chair of Data Science in Earth Observation, Technical University of Munich, Germany}\\}
\date{}
\maketitle

\begin{abstract}
High-resolution satellite images can provide abundant, detailed spatial information for land cover classification, which is particularly important for studying the complicated built environment. However, due to the complex land cover patterns, the costly training sample collections, and the severe distribution shifts of satellite imageries caused by, e.g., geographical differences or acquisition conditions, few studies have applied high-resolution images to land cover mapping in detailed categories at large scale. To fill this gap, we present a large-scale land cover dataset, \emph{Five-Billion-Pixels}. It contains more than \emph{5 billion} labeled pixels of $150$ high-resolution Gaofen-2 ($4$ m) satellite images, annotated in a 24-category system covering \emph{artificial-constructed}, \emph{agricultural}, and \emph{natural} classes. In addition, we propose a deep-learning-based unsupervised domain adaptation approach that can transfer classification models trained on labeled dataset (referred to as the \emph{source domain}) to unlabeled data (referred to as the \emph{target domain}) for large-scale land cover mapping. Specifically, we introduce an end-to-end Siamese network employing dynamic pseudo-label assignment and class balancing strategy to perform adaptive domain joint learning. To validate the generalizability of our dataset and the proposed approach across different sensors and different geographical regions, we carry out land cover mapping on five megacities in China and six cities in other five Asian countries severally using: PlanetScope ($3$ m), Gaofen-1 ($8$ m), and Sentinel-2 ($10$ m) satellite images. Over a total study area of 60,000 km$^{2}$, the experiments show promising results even though the input images are entirely unlabeled. The proposed approach, trained with the \emph{Five-Billion-Pixels} dataset, enables high-quality and detailed land cover mapping across the whole country of China and some other Asian countries at meter-resolution.
\end{abstract}

\section{Introduction}
\subsection{Motivation}
Land cover information is crucial for various research fields involving environment science, climate monitoring, food security, urban planning, disaster management, and ecosystem protection \cite{objectCNN2}. With the continuous development of technology and the economy, human activities have an increasing impact on both urban and natural environments \cite{Global,2017multi}. There is, therefore, an urgent need for timely and reliable large-scale land cover information to guide the construction of human settlements and to mitigate the negative environmental changes.

Over the past few decades, extensive studies have been devoted to large-scale land cover mapping using low-/medium-spatial resolution remote sensing images \cite{lowResolution1,lowResolution2,2017breaking,2019stable,urban,corine}, e.g., Moderate Resolution Imaging Spectroradiometer (MODIS) \cite{lowMODIS}, Landsat Thematic Mapper (TM) \cite{mediumLandsat,202130}, Enhanced Thematic Mapper+ (ETM+) \cite{mediumETM+} satellite imageries, and remarkable achievements have been accumulated. However, due to the lack of spatial information, these images are insufficient to distinguish heterogeneous land cover categories, especially for the categories mainly distributed in the built environment, such as buildings, traffic infrastructures, artificial water areas, and urban green spaces. Recently, based on Sentinel satellite imagery, European Space Agency (ESA) and Google have released global 10 m land cover mapping projects, World Cover \cite{esaLC} and Dynamic World \cite{googleLC}, respectively. Although they are highly accurate and real-time, they cover only basic land cover categories (11 and 9 classes) and have limited ability to depict urban environments.

Owing to the advances in satellite technology, remote sensing images with higher spatial resolution are becoming increasingly available. Compared with low-/medium-spatial resolution images, they provide richer texture, shape, and spatial distribution information of ground objects, which contribute significantly to detailed mapping in high heterogeneous areas, e.g., densely populated megacities. But at the same time, the detailed information brings about much more complicated land structures and patterns \cite{challenge}, which leads to great challenges in land cover classification with high-resolution images. Furthermore, due to the narrow swath of high-resolution images and the problem of cloud obscuration, it is often necessary to jointly use numerous images captured at different times and positions by the same sensor, or even by multiple satellite sensors, to mosaic large-scale land cover maps \cite{threshold1}. The attendant problem is that diverse imaging conditions will lead to shifts in feature distributions of ground objects, which causes the classification method optimal for certain annotated images (referred to as the \emph{source domain}) to drop drastically in performance on newly acquired images (referred to as the \emph{target domain}) \cite{DomainAdaptation,GID}. The above factors render it difficult for high-resolution satellite images to be employed practically for large-scale land cover mapping applications.

\subsection{Related work}
In recent years, tremendous effort has been dedicated to this challenging task. In the early stage, spectral and spectral-spatial features were widely utilized to identify land cover categories based on pixel- or object-spatial units \cite{spectralSpatial1,spectralSpatial2,spectralSpatial3,2019automatic}. Nevertheless, restricted, hand-crafted rules cannot fully define and represent the complicated land structures or patterns in high-resolution images \cite{challenge}. To address this problem, deep learning has attracted broad attention in the remote sensing community. Deep Convolutional Neural Networks (DCNNs) are able to adaptively approximate the relationship between image information and land information through multi-layer transformations \cite{deep1}. Thus, compared with conventional land cover classification methods, deep models can accurately characterize complex contextual information contained in high-resolution images \cite{GID,objectCNN1,objectCNN2,objectCNN3,discrepancy2}. Although deep models have reported great superiorities in many remote sensing issues \cite{deep1,deep2, zhu2021}, their performance strongly relies on the quality and quantity of training data \cite{DeepLearning,2017aid,2021object}, resulting in two main problems in applying them to real-world land cover mapping:

\begin{itemize}
\item[-]\emph{The application gap caused by limited representativity of land cover datasets}: Deep learning is a data-driven approach, and its potential for practical land cover mapping depends heavily on whether the training data fully reflects the distribution of real-world ground objects. An insufficient amount of data may lead to overfitting of the model, insufficient data diversity may lead to low generalization capability of the model, and an incomplete category system will make the model unable to meet the actual mapping requirements \cite{dataset,earthnets}. 
\item[-]\emph{The inadequate generalizability of deep models over different data domains}: Even if a practicable deep model is already trained on a well-annotated dataset, it may not be valid for other geographical areas or sensors because of the feature distribution shifts between the source and target domains \cite{DomainAdaptation,GID,huang2023}. To adapt this deep model to large-scale land cover mapping, an intuitive way is annotating sufficient samples for the target domain and performing model retraining. However, dense annotation for each newly captured image is not realistic.
\end{itemize}

To alleviate the first problem, a number of densely labeled land cover datasets with sub-meter to meter-spatial resolutions ($0.05$-$10$ m) have been released and contributed substantially to land cover classification research. But most of them have geographical coverage areas below $10$ km$^{2}$ and are located in concentrated regions, such as ISPRS Potsdam \cite{ISPRS2D}, ISPRS Vaihingen \cite{ISPRS2D}, Zurich Summer \cite{Zurich}, RIT-18 \cite{RIT-18}, and Zeebruges \cite{Zeebruges}. Existing large-scale datasets, with coverage areas more than $1000$ km$^{2}$ and wide geographical distributions, are typically annotated with about 10 classes, and do not contain detailed urban functional categories, including SpaceNet \cite{SpaceNet}, DeepGlobe \cite{DeepGlobe},  MiniFrance \cite{2021MiniFrance}, Gaofen Image Dataset (GID) \cite{GID,GID15}, and LandCoverNet \cite{LandCoverNet}. Although these large-scale datasets possess adequate data amount and data diversity, their incomplete land cover category systems prevent them from fully bridging the gap between algorithmic research and real-world applications.

To solve the second problem, unsupervised domain adaptation (UDA) has been commonly considered by recent remote sensing literature \cite{GID,discrepancy1,discrepancy2,adversarial1,adversarial2,2022multitarget}. UDA aims to adapt models trained on the source domain to the target domain without supervised information \cite{DomainAdaptation}. Two major types of deep-learning-based UDA have been studied: discrepancy-based and adversarial-based methods. Discrepancy-based methods minimize the discrepancy criteria between the source and target domains to reduce the distance of their distributions \cite{discrepancy1,discrepancy2}. The discrepancy criteria are implemented in the form of manually designed loss functions, such as Correlation Alignment (CORAL) \cite{coral} and Maximum Mean Discrepancy (MMD) \cite{MMD}. In contrast, adversarial-based methods, such as the domain-adversarial neural network (DANN) \cite{DANN} and Adversarial Discriminative Domain Adaptation (ADDA) \cite{ADDA}, do not require manually designed criteria for domain matching. They instead learn criteria by simultaneously training a feature generator and a domain discriminator, which attempt to extract indistinguishable features for both domains and distinguish the features of different domains, respectively \cite{adversarial1,adversarial2}.

The essential idea of these two types of UDA methods is to align the feature distributions of the source and target domains \cite{coral,DANN,loveda}. However, this idea works on a key assumption that it is possible to find an appropriate match for the two distributions, while real-world situations are often not that ideal. First, for large-scale land cover mapping, both domains may contain images from diverse imaging conditions, resulting in a widely dispersed feature space within each domain. In this context, a rigid alignment of two dispersive domains may further accumulate intra-domain variance \cite{negativeAdapt,proportion2}. Secondly, class imbalance is prevalent in the actual land cover patterns. The most common category may cover an area hundreds of times larger than some other categories. During a global alignment, a few common categories may, thus, pull the entire domain toward their optimal distribution, causing negative adaptation to other categories. 

To improve the performance of UDA, recent works \cite{threshold1,proportion1,lovecs} have combined distribution alignment with pseudo-labeling. The main idea of pseudo-labeling is to select valuable samples from the target domain  for model training according to the predicted classification confidence. However, the quality of pseudo-labels depends on the way they are selected. The current approaches are empirically choosing a confidence threshold \cite{threshold1,threshold2} or setting a fixed proportion for sample collection \cite{GID,proportion1,proportion2,lovecs}, which is hard to guarantee the accuracy of pseudo-labels assigned by the prediction model to its unknown domain. Furthermore, the model always tends to select more easy samples, which may exacerbate the category imbalance. These issues make it difficult for existing UDA methods to satisfy the demands of large-scale land cover mapping applications.

\subsection{Contribution of this paper}
To address the above-mentioned problems, in this paper, we propose the \emph{Five-Billion-Pixels} dataset, which extends the land cover dataset GID \cite{GID,GID15}. Instead of 5/15 basic categories of GID, it contains more than \emph{5 billion} labeled pixels of $150$ high-resolution Gaofen-2 (GF-2) satellite images annotated in a more complete category system, consisting of 24 land use and land cover classes. Then, we propose a UDA approach for practical large-scale land cover mapping. Instead of the domain alignment strategy, our approach lets the deep model already defined on the source domain gradually and adaptively learn the distribution of the target domain. Concretely, we introduce a Siamese network \cite{siamese} with two branches that separately generate feature maps for images from the source and target domain. The branches share the same fully convolutional architecture and the same parameters pre-trained on \emph{Five-Billion-Pixels}. In the target domain branch, information entropy of the feature maps is treated as the indicator to select image pixels with high confidence, and category predictions on the selected pixels are considered as pseudo-labels. These pseudo-labels are then used to construct a joint classification loss with the source domain branch. To trade off the adaptation over two domains, the number of pixels assigned with pseudo-labels is dynamically changed with training iterations. To prevent over-adaptation to some common categories, the joint classification loss is weighted according to the class distribution in the source domain.

The main contributions of this paper are as follows:

\begin{itemize}
\item[-]We present a large-scale land cover classification dataset, \emph{Five-Billion-Pixels}. It has the spatial resolution of \emph{4} m, covers areas over \emph{50,000} km$^{2}$ in China, and contains more than \emph{5 billion} labeled pixels. Its category system covers \emph{artificial-constructed}, \emph{agricultural}, and \emph{natural} classes, which well-reflects the distribution of real-world ground objects and can widely benefit land-cover-related studies.
\item[-]We propose a deep-learning-based UDA approach for large-scale land cover mapping. It avoids changing the domain distributions in a rigid way but softly corrects the domain shifts according to the knowledge learned in the source domain. The negative adaptation caused by intra-domain diversity and class imbalance can, therefore, be mitigated by our approach even under very complicated practical situations.
\item[-]We carry out land cover mapping on five megacities in China and six cities in other five Asian countries severally using unlabeled PlanetScope ($3$ m), Gaofen-1 ($8$ m), and Sentinel-2 ($10$ m) satellite images. Encouraging experimental results are achieved over a total area of 60,000 km$^{2}$, demonstrating the potential of the proposed dataset and approach for high-quality, detailed land cover mapping across the whole country of China and some other Asian countries at meter-resolution.
\end{itemize}

\begin{figure*}[htb!]
\centering
\includegraphics[width=0.7\textwidth]
{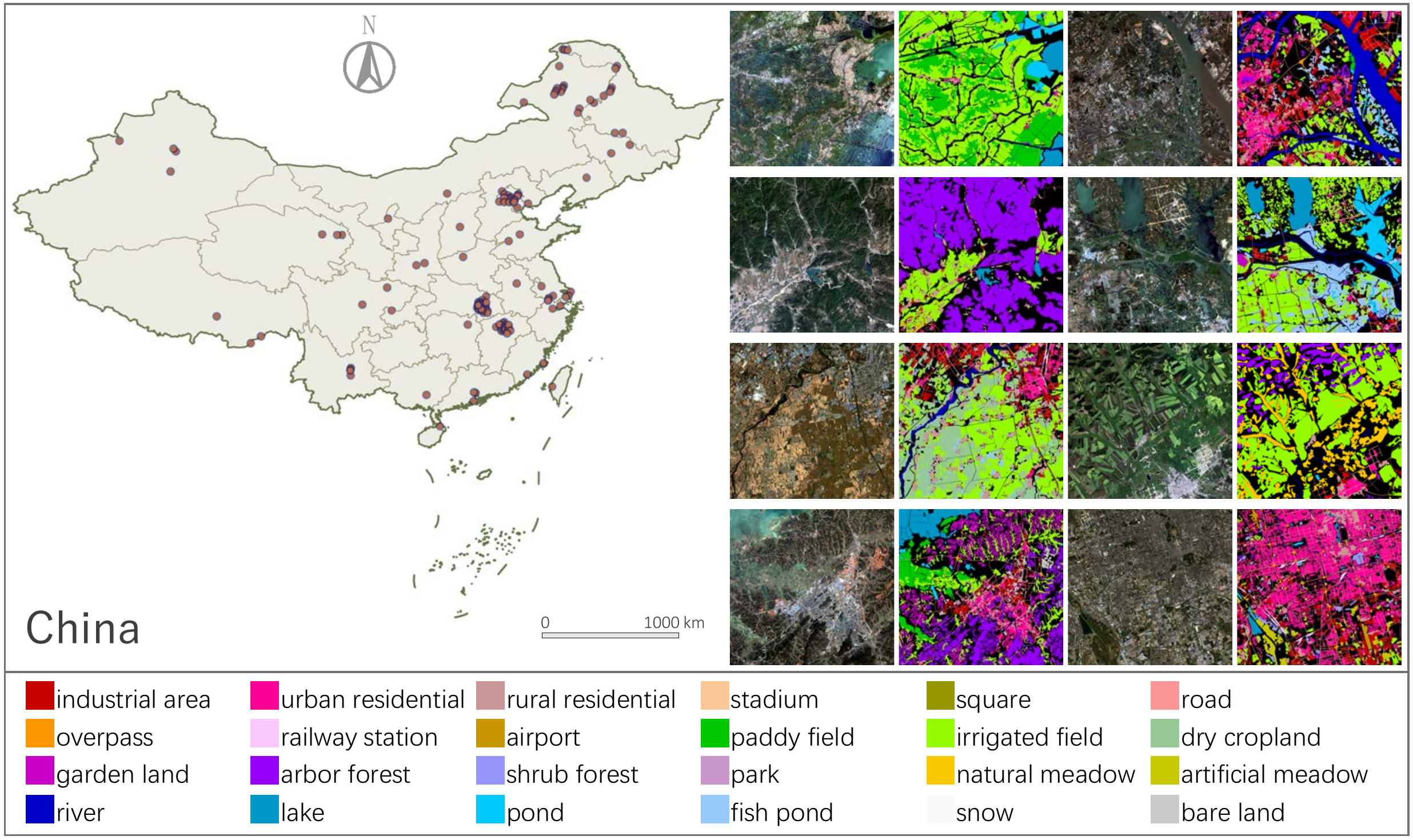}
\caption{Left: Distribution of 150 images of \emph{Five-Billion-Pixels}. Right: Examples of GF-2 images and their corresponding label maps, where black indicates unlabeled areas. The category system covers artificial-constructed, agricultural, and natural classes.}
\label{figure:distribution}
\end{figure*}

\section{Study data}
To reduce the gap between high-resolution land cover datasets and real-world application requirements, we reorganize and augment the category system of the land cover dataset GID. GID is available in versions with 5/15 classes; interested readers can refer to \cite{GID} and \cite{GID15}. Our new dataset, named \emph{Five-Billion-Pixels}, consists of $150$ GF-2 satellite images annotated in a more complete category system (see Fig. \ref{figure:distribution}). It has the advantages of rich categories, large coverage, wide distribution, and high-spatial resolution of $4$ m.

For the case study of large-scale land cover mapping, we perform land cover classification for five megacities in China and six cities in other five Asian countries using three data sources with diverse spatial resolutions. Concretely, for Chinese megacities, PlanetScope (PS) satellite images are used for Chengdu and Shanghai, Gaofen-1 (GF-1) satellite images are used for Wuhan, and Sentinel-2 (ST-2) satellite images are used for Beijing and Guangzhou. And for other Asian cities: Bangkok, Thailand; Delhi, India; Naypyidaw, Myanmar; Seoul, South Korea; Tokyo, Japan; and Yangon, Myanmar, ST-2 satellite images are used. The Chinese megacities cover a total geographical area of 53,088 km$^{2}$ and are separately located in the eastern, western, northern, southern, and central regions of China. And the other six Asian cities are located in South, Southeast, and East Asia, respectively.

The \emph{Five-Billion-Pixels} dataset is introduced in Section \ref{sec:datagid24}, and the study areas with their data sources are introduced in Section \ref{sec:datamulti}.

\subsection{Five-Billion-Pixels}
\label{sec:datagid24}

\subsubsection{Gaofen-2 imagery}
GF-2 is the second satellite of the High-Definition Earth Observation System (HDEOS) promoted by China National Space Administration (CNSA). It is equipped with two panchromatic and multispectral (PMS) sensors, providing a combined swath of $45$ km. The effective spatial resolution of the sensors is $1$ m panchromatic (pan)/$4$ m multispectral (MS). The MS images we used to construct \emph{Five-Billion-Pixels} possess a spectral range of blue ($0.45$-$0.52$ $\upmu$m), green ($0.52$-$0.59$ $\upmu$m), red ($0.63$-$0.69$ $\upmu$m), and near-infrared ($0.77$-$0.89$ $\upmu$m), with an image resolution of $6800\times 7200$ pixels. Owing to the combination of high-resolution and wide swath, GF-2 allows the observation of detailed land information over large geographical areas.

\subsubsection{Creation of Five-Billion-Pixels}
The creation of \emph{Five-Billion-Pixels} fully relied on human manual annotation. To maximize label consistency and minimize human error, the annotation process contains four phases: coarse labeling, fine labeling, fine checking, and spot checking. 

\begin{figure*}[htb!]
\centering
\includegraphics[width=0.7\textwidth]
{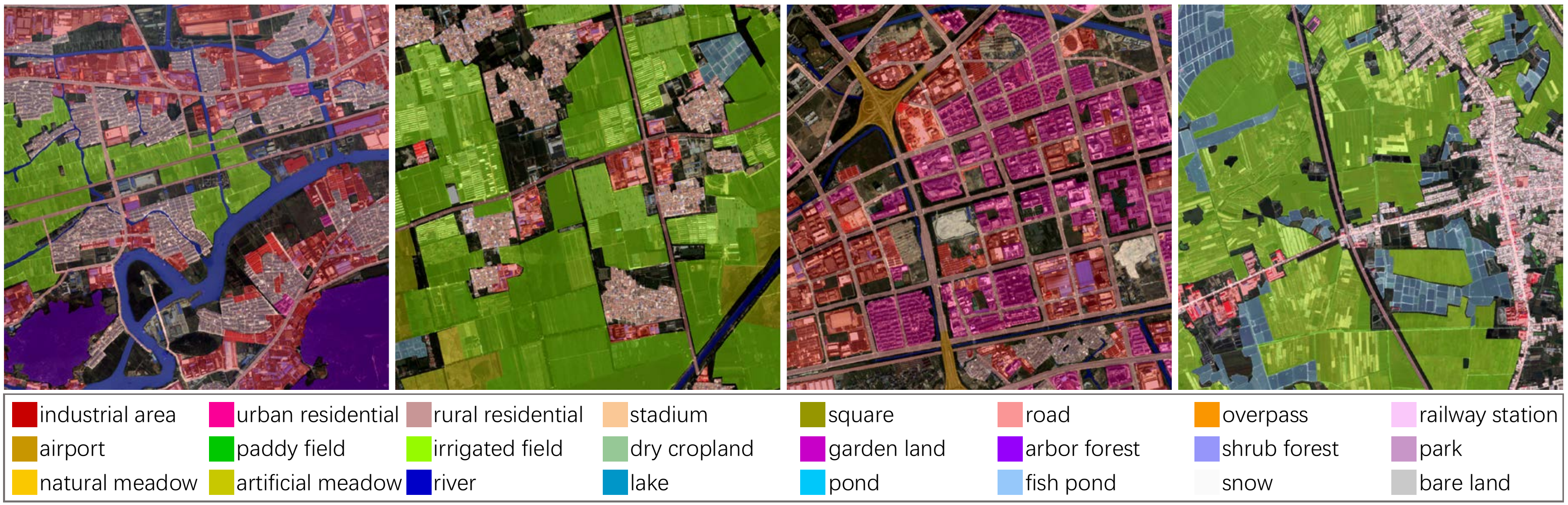}
\caption{Examples of annotation details. Miscellaneous or unclear areas that are extremely difficult to annotate are considered as unlabeled. The categories of labeled pixels are double-checked and ensured to be correct.}
\label{figure:detailedlabel}
\end{figure*}

First, the category system of \emph{Five-Billion-Pixels} is determined with reference to Chinese Land Use Classification Criteria (\emph{GB/T 21010-2017}), and the classes are adjusted based on the recognizability of 4 m-resolution optical remote sensing images. During the coarse labeling process, the interpretation experts roughly delineate regions belonging to different classes on each GF-2 image according to the category system. For uncertain areas, Google Earth and Google Map with corresponding geographic coordinates are considered as references. These rough annotations are then passed to the labeling crew for fine labeling. The labeling crew uses the lasso tool in Adobe Photoshop software to frame the ground objects so that the edge of the label map and the edge of the ground objects can be strictly coincident; some details of annotations are presented in Fig. \ref{figure:detailedlabel}. Fine checking consists of two rounds, check of categories and check of edges. The interpretation experts carefully collate each area of each label map and mark the inaccurate categories or edges, which are then passed to the labeling crew for correction. The final spot checking is to slice GF-2 images and corresponding label maps into $500\times 500$-pixel patch pairs, at which scale it is easier to find errors, and present the patch pairs randomly to the interpretation experts for inspection. The inspection results are then given to the labeling crew for modification, and the interpretation experts conduct the next round of spot checking on the revised results. In the final round of spot checking, 10\% samples of \emph{Five-Billion-Pixels} are examined and no obvious errors are observed.

\subsubsection{Properties of Five-Billion-Pixels}
\textbf{Rich Categories}: The category system of \emph{Five-Billion-Pixels} concretely includes: \emph{industrial area}, \emph{urban residential}, \emph{rural residential}, \emph{stadium}, \emph{square}, \emph{road}, \emph{overpass}, \emph{railway station}, \emph{airport}, \emph{paddy field}, \emph{irrigated field}, \emph{dry cropland}, \emph{garden land}, \emph{arbor forest}, \emph{shrub forest}, \emph{park}, \emph{natural meadow}, \emph{artificial meadow}, \emph{river}, \emph{lake}, \emph{pond}, \emph{fish pond}, \emph{snow}, \emph{bare land}. Miscellaneous or unclear areas that are extremely difficult to annotate are considered as unlabeled. This category system covers artificial-constructed, agricultural, and natural classes, more closely resembling the distributions of ground objects in the real world. Notably, the category system contains a number of land use classes subdivided from land cover classes in \emph{GB/T 21010-2017}, including: \emph{stadium} and \emph{square} from public service land; \emph{road}, \emph{overpass}, \emph{railway station}, and \emph{airport} from transportation land; \emph{park} and \emph{artificial meadow} from artificial non-agricultural vegetated areas. This is designed to make full use of the spatial information of high-resolution images and to enrich the application scenarios of urban environmental analysis. As \emph{Five-Billion-Pixels} are mainly collected from human activity areas (cities, villages, cultivated lands, and mountainous areas around cities), the category system covers all land categories except \emph{mangroves}, \emph{tundra} and \emph{permanent ice}. The percentage of pixels belonging to each category among all labeled pixels is listed in Table \ref{table:ctgrratio}.

\begin{table*}[htb!]
\centering
\caption{The percentage of the number of pixels belonging to each category. The abbreviations for categories are defined as: Indu - \emph{industrial area}, Urba - \emph{urban residential}, Rura - \emph{rural residential}, Stad - \emph{stadium}, Squa - \emph{square}, Over - \emph{overpass}, Rail - \emph{railway station}, Airp - \emph{airport}, Padd - \emph{paddy field}, Irri - \emph{irrigated field}, Dryc - \emph{dry cropland}, Gard - \emph{garden land}, Arbo - \emph{arbor forest}, Shru - \emph{shrub forest}, Natu - \emph{natural meadow}, Arti - \emph{artificial meadow}, Rive - \emph{river}, Fish - \emph{fish pond}, Bare - \emph{bare land}. The category proportion is not deliberately controlled, but it is labeled according to the distribution of real-world ground objects. As can be seen, the category distribution of \emph{Five-Billion-Pixels} is quite imbalanced.}
\vspace{3mm}
\resizebox{0.8\textwidth}{!}{
\begin{tabular}{lllllllllllll} 
\hline
\textbf{Category}     & Indu & Urba & Rura & Stad & Squa & Road & Over & Rail & Airp & Padd & Irri & Dryc \\
\textbf{Percent (\%)} & 3.57 & 5.60 & 4.39 & 0.02 & 0.02 & 3.57 & 0.23 & 0.08 & 0.09 & 2.40 &37.26 & 6.65 \\ 
\hline
\textbf{Category}     & Gard & Arbo & Shru & Park & Natu & Arti & Rive & Lake & Pond & Fish & Snow & Bare \\
\textbf{Percent (\%)} & 0.91 & 8.05 & 3.80 & 0.05 & 1.65 & 0.36 & 5.08 & 9.87 & 1.03 & 1.12 & 0.03 & 4.16 \\
\hline
\end{tabular}}
\label{table:ctgrratio}
\end{table*}

\textbf{Large Coverage}: The 150 GF-2 satellite images contained in \emph{Five-Billion-Pixels} have a total geographical coverage of over 50,000 km$^{2}$. On this basis, more than $5$ billion pixels are carefully annotated, which can provide abundant samples for advancing research in data-driven methodologies.

\textbf{Wide Distribution}: The image source of \emph{Five-Billion-Pixels} is collected from more than 60 dispersed administrative districts in China, as Fig. \ref{figure:distribution} shows. Due to the wide geographical distribution, \emph{Five-Billion-Pixels} can reflect the variation of landscapes with different climate, altitude, and geology.

\subsection{Study areas and data sources}
\label{sec:datamulti}

\subsubsection{Chinese megacities}

\begin{figure*}[htb!]
\centering
\includegraphics[width=0.7\textwidth]
{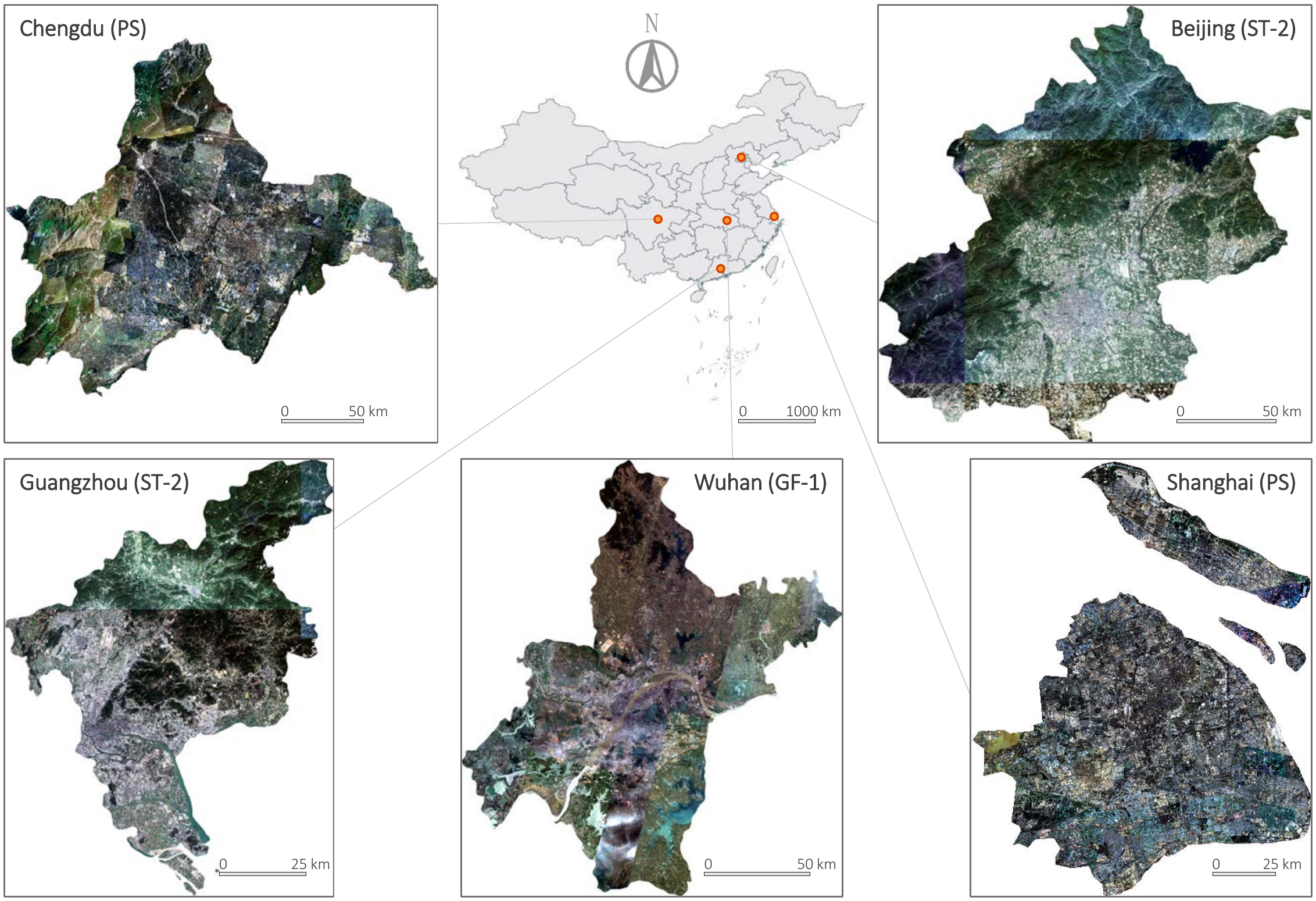}
\caption{Chinese megacities and their locations.}
\label{figure:studychina}
\end{figure*}

We select five Chinese megacities with diverse geographical environments, development degrees, and city structures as our study areas: Beijing, Chengdu, Guangzhou, Shanghai, and Wuhan.

Real-time, large-scale land cover mapping may require the joint use of images captured by multiple sensors; hence, the adaptation of the classification approach to diverse sensors is of great significance. Based on this consideration, we construct our study areas with imageries from three different sensors, as shown in Fig. \ref{figure:studychina}.

Specifically, the study data of Beijing are mosaicked from 9 ST-2 images acquired between November 8, 2020, and October 21, 2021. The data of Chengdu are mosaicked from 205 PS images captured between January 13, 2019, and December 31, 2019. The data of Guangzhou are mosaicked from 3 ST-2 images collected between February 18, 2021, and October 26, 2021. The data of Shanghai are mosaicked from 149 PS images obtained between April 1, 2019, and December 13, 2019. And the data of Wuhan are mosaicked from 22 GF-1 images taken between March 28, 2016, and July 25, 2016.

\begin{figure*}[htb!]
\centering
\includegraphics[width=0.7\textwidth]
{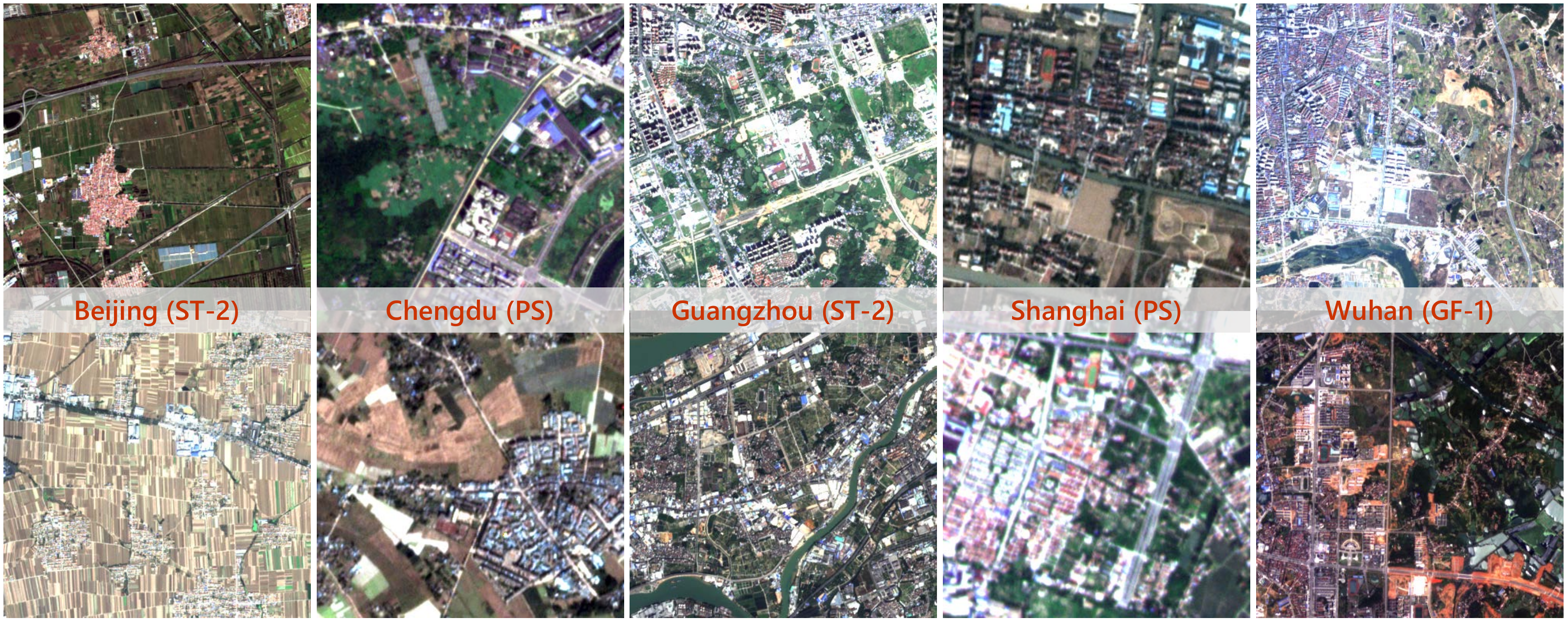}
\caption{Each column indicates images from the same city. There are  obvious spectral shifts even between images within the same city due to differences in imaging illumination and season. In addition, spatial resolutions of different data sources are distinctly diverse. These heterogeneities pose huge challenges to land cover mapping.}
\label{figure:multisource}
\end{figure*}

There is great heterogeneity in images acquired by different sensors. And, due to the impact of cloud obscuration, swath width, and revisit period, it is necessary to utilize images captured in different seasons and lighting conditions to mosaic the complete image map for each city. As a result, not only are there significant differences between data sources, but there are also distribution shifts between images within each city, as displayed in Fig. \ref{figure:multisource}.

\subsubsection{Additional Asian cities}
\begin{figure*}[htb!]
\centering
\includegraphics[width=0.7\textwidth]
{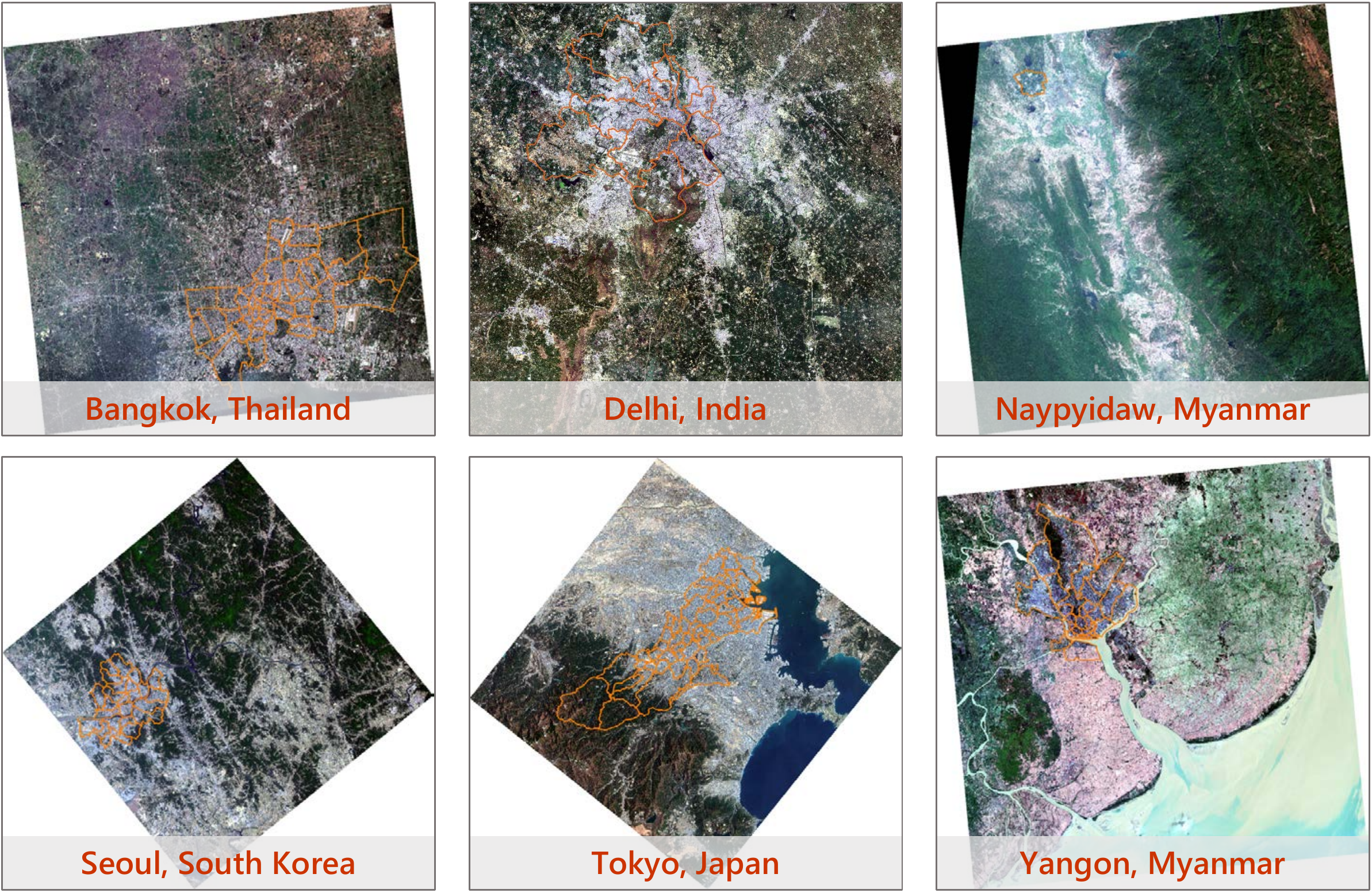}
\caption{Administrative areas of six Asian cities on ST-2 images.}
\label{figure:studyasia}
\end{figure*}

To verify the applicability of our approach to different regions in the world, we select six cities in five Asian countries as study areas: Bangkok, Thailand; Delhi, India; Naypyidaw, Myanmar; Seoul, South Korea; Tokyo, Japan; and Yangon, Myanmar, as shown in Fig. \ref{figure:studyasia}.

The data source used for land cover mapping of these Asian cities is ST-2 satellite imagery. And the images for Bangkok, Delhi, Naypyidaw, Seoul, Tokyo, and Yangon are separately captured on January 6, 2022; March 5, 2022; February 1, 2022; May 17, 2022; February 21, 2021; and January 7, 2022.

As can be seen, although we treat ``city'' as study subject, our experimental areas actually contain other types of landscapes besides built-up. For Chinese megacities, their administrative regions cover large agricultural land and forests, and for the additional Asian cities, we classify the entire images, i.e. including the surrounding areas outside the administrative regions. Therefore, these study areas can test the performance of classification approaches for urban, countryside, agricultural, and mountainous scenes.

\subsubsection{Data sources}
\textbf{PlanetScope}: PS is a satellite constellation of about 130 individual CubeSats operated by American Planet Lab. Its sensors capture MS images in blue ($0.46$-$0.52$ $\upmu$m), green ($0.50$-$0.59$ $\upmu$m), red ($0.59$-$0.67$ $\upmu$m), and near-infrared ($0.78$-$0.86$ $\upmu$m) bands with a spatial resolution of $3.7$-$4.1$ m, which is resampled to approximately $3$ m at data release.

\textbf{Gaofen-1}: GF-1 is the first satellite of HDEOS proposed by China. It is configured with two PMS, providing a spatial resolution of $2$ m pan/$8$ m MS and a combined swath of over $60$ km. The MS images used in our study cover the spectral range of blue ($0.45$-$0.52$ $\upmu$m), green ($0.52$-$0.59$ $\upmu$m), red ($0.63$-$0.69$ $\upmu$m), and near-infrared ($0.77$-$0.89$ $\upmu$m).

\textbf{Sentinel-2}: ST-2 is an Earth observation mission from the European Union's Copernicus Programme. It currently comprises a constellation with two satellites, Sentinel-2A and Sentinel-2B, offering 13 spectral bands and a field of view of $290$ km. Blue (central wavelength $0.49$ $\upmu$m), green (central wavelength $0.56$ $\upmu$m), red (central wavelength $0.66$ $\upmu$m), and near-infrared (central wavelength $0.83$ $\upmu$m) bands with a resolution of $10$ m are used in our study. Because of the free, open data policy and the advantages in spatial and spectral resolution, ST-2 is one of the most commonly used data sources for recent land cover mapping studies \cite{st2Mapping1,st2Mapping2,st2Mapping3}.

\subsubsection{Test areas}
\begin{figure*}[htb!]
\centering
\includegraphics[width=0.7\textwidth]
{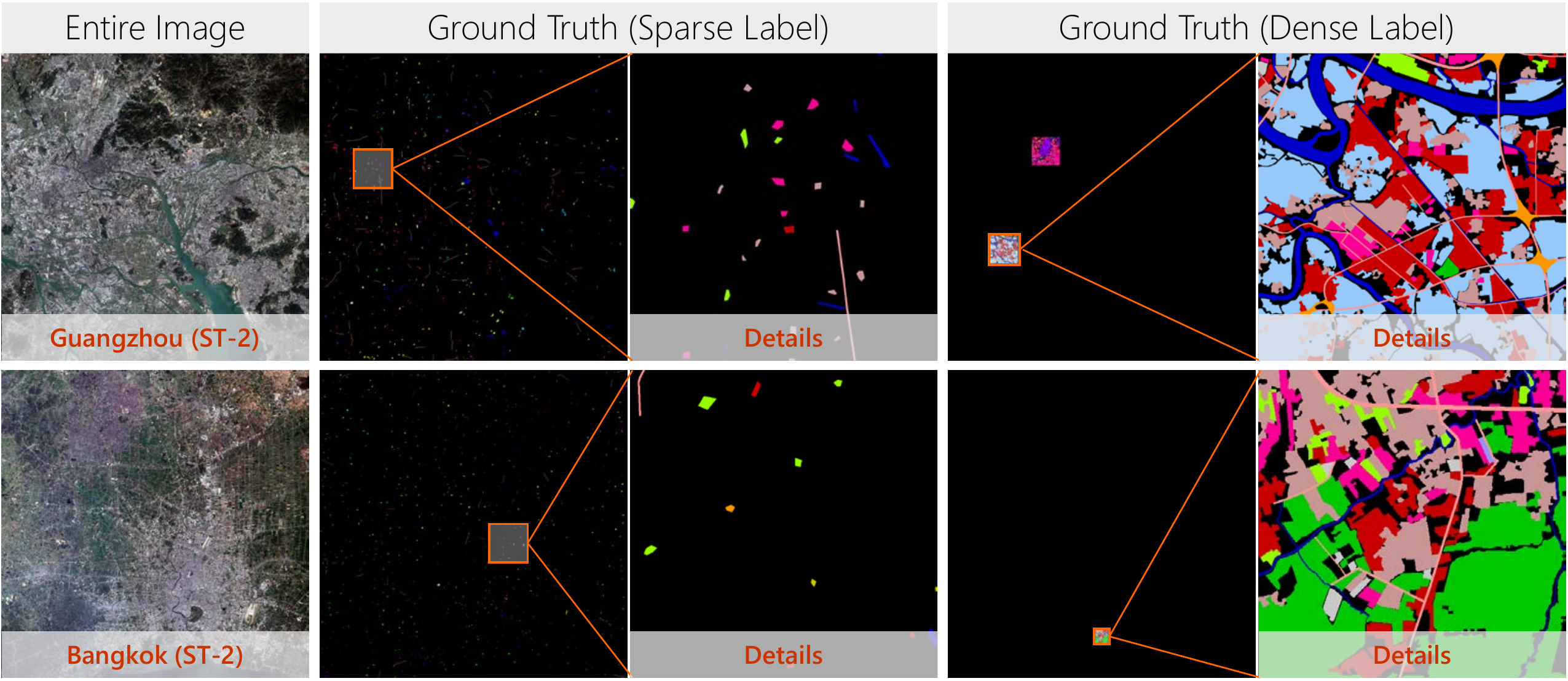}
\caption{Two annotation strategies for quantitative evaluation. Sparse label: small polygons are evenly labeled throughout the entire image. Dense label: sub-regions with sizes of $1000\times 1000$ and $500\times 500$ pixels are labeled for Chinese megacities and the additional Asian cities, respectively.}
\label{figure:studytest}
\end{figure*}

Because the study areas are overly large, it is impossible to densely annotate each test image for quantitative evaluation. We therefore adopt two annotation strategies, sparse labeling and dense labeling. Concretely, sparse labeling is to evenly annotate small polygons on each image of each city, and dense labeling is to densely annotate sub-regions for each city. For dense labeling strategy, each Chinese megacity is labeled with two sub-regions of $1000\times 1000$ pixels, and each additional Asian city is labeled with a sub-region of $500\times 500$ pixels since their administrative districts are smaller, as illustrated in Fig. \ref{figure:studytest}.

\begin{table*}[htb!]
\centering
\caption{The percentage of the number of pixels belonging to each category in Chinese megacity test areas. Sparse label contains $5.21\times 10^{7}$ pixels and dense label covers $7.77\times 10^{6}$ pixels.}
\vspace{3mm}
\resizebox{0.75\textwidth}{!}{
\begin{tabular}{lllllllllllll} 
\hline
\textbf{Category}     & Indu & Urba & Rura & Stad & Squa & Road & Over & Rail & Airp & Padd & Irri & Dryc \\
\textbf{Sparse (\%)}  & 7.04 & 5.49 & 2.93 & 1.02 & 0.50 &11.25 & 2.66 & 2.25 & 1.96 & 1.89 &18.11 & 1.50 \\
\textbf{Dense (\%)}   & 9.36 &35.00 & 8.51 & 0.09 & 0.07 & 6.80 & 0.50 & 0.15 & 0.28 & 8.53 &13.82 & 0.07 \\
\hline
\textbf{Category}     & Gard & Arbo & Shru & Park & Natu & Arti & Rive & Lake & Pond & Fish & Snow & Bare \\
\textbf{Sparse (\%)}  & 3.44 & 6.45 & 0.45 & 2.28 & 1.41 & 0.65 &14.47 & 7.47 & 1.34 & 1.64 & 0.61 & 3.16 \\
\textbf{Dense (\%)}   & 0.34 & 2.58 & 0.08 & 1.23 & 0.07 & 0.37 & 5.64 & 0.72 & 0.53 & 4.72 & 0    & 0.56 \\
\hline
\end{tabular}}
\label{table:ratiomegacity}
\end{table*}

\begin{table*}[htb!]
\centering
\caption{The percentage of each category in the additional Asian city test areas. Sparse label includes $2.40\times 10^{6}$ pixels and dense label covers $1.29\times 10^{6}$ pixels.}
\vspace{3mm}
\resizebox{0.75\textwidth}{!}{
\begin{tabular}{lllllllllllll} 
\hline
\textbf{Category}     & Indu & Urba & Rura & Stad & Squa & Road & Over & Rail & Airp & Padd & Irri & Dryc \\
\textbf{Sparse (\%)}  & 5.81 & 3.89 &11.56 & 0.29 & 0.35 & 4.75 & 0.94 & 0.69 & 1.20 & 5.72 &31.39 & 0.19 \\
\textbf{Dense (\%)}   & 5.47 & 6.91 &44.32 & 0    & 0    & 5.21 & 0.42 & 0.10 & 0.60 & 6.92 &20.68 & 0    \\
\hline
\textbf{Category}     & Gard & Arbo & Shru & Park & Natu & Arti & Rive & Lake & Pond & Fish & Snow & Bare \\
\textbf{Sparse (\%)}  & 2.23 &12.05 & 0.97 & 0.39 & 0.07 & 0.33 & 7.46 & 4.56 & 2.39 & 0.25 & 0.24 & 2.26 \\
\textbf{Dense (\%)}   & 0.25 & 5.16 & 0.27 & 0    & 0    & 0.05 & 3.15 & 0    & 0.17 & 0    & 0    & 0.32 \\
\hline
\end{tabular}}
\label{table:ratioasiacity}
\end{table*}

In total, the five Chinese megacities are sparsely labeled with $5.21\times 10^{7}$ pixels and densely labeled with $7.77\times 10^{6}$ pixels, and the percentage of each category in test areas is displayed in Table \ref{table:ratiomegacity}.

The additional Asian cities are sparsely labeled with a total of $2.40\times 10^{6}$ pixels and densely labeled with a total of $1.29\times 10^{6}$ pixels, and the percentage of each category in test areas is listed in Table \ref{table:ratioasiacity}.

Sparse label covers all categories and can be used to evaluate the performance of classification on the entire images. And dense label can be utilized to verify the fineness of the classification results in local areas. Note that these annotations are only used for accuracy assessment and not for model training.

\section{Methodology}
To adapt the knowledge learned from the labeled dataset to land cover mapping of large-scale areas, we propose a UDA approach that can softly correct the domain shifts by adaptively learning the distribution of unlabeled data. We refer to two domains, the source domain as $\textbf{\emph{D}}_{\textbf{\emph{S}}}$ and the target domain as $\textbf{\emph{D}}_{\textbf{\emph{T}}}$, representing the \emph{Five-Billion-Pixels} dataset and the unannotated images used for land cover mapping, respectively.

First, we utilize $\textbf{\emph{D}}_{\textbf{\emph{S}}}$ to pre-train a semantic segmentation model, which is presented in Section \ref{sec:segment}. Subsequently, we take the pre-trained semantic segmentation model as the backbone to construct a Siamese network, of which the two identical branches separately process images for $\textbf{\emph{D}}_{\textbf{\emph{S}}}$ and $\textbf{\emph{D}}_{\textbf{\emph{T}}}$. In the $\textbf{\emph{D}}_{\textbf{\emph{T}}}$ branch, a fraction of image pixels with high confidence is selected and then used to perform domain joint learning with the $\textbf{\emph{D}}_{\textbf{\emph{S}}}$ branch, which is described in Section \ref{sec:adaptat}.

\begin{figure*}[htb!]
\centering
\includegraphics[width=0.7\textwidth]
{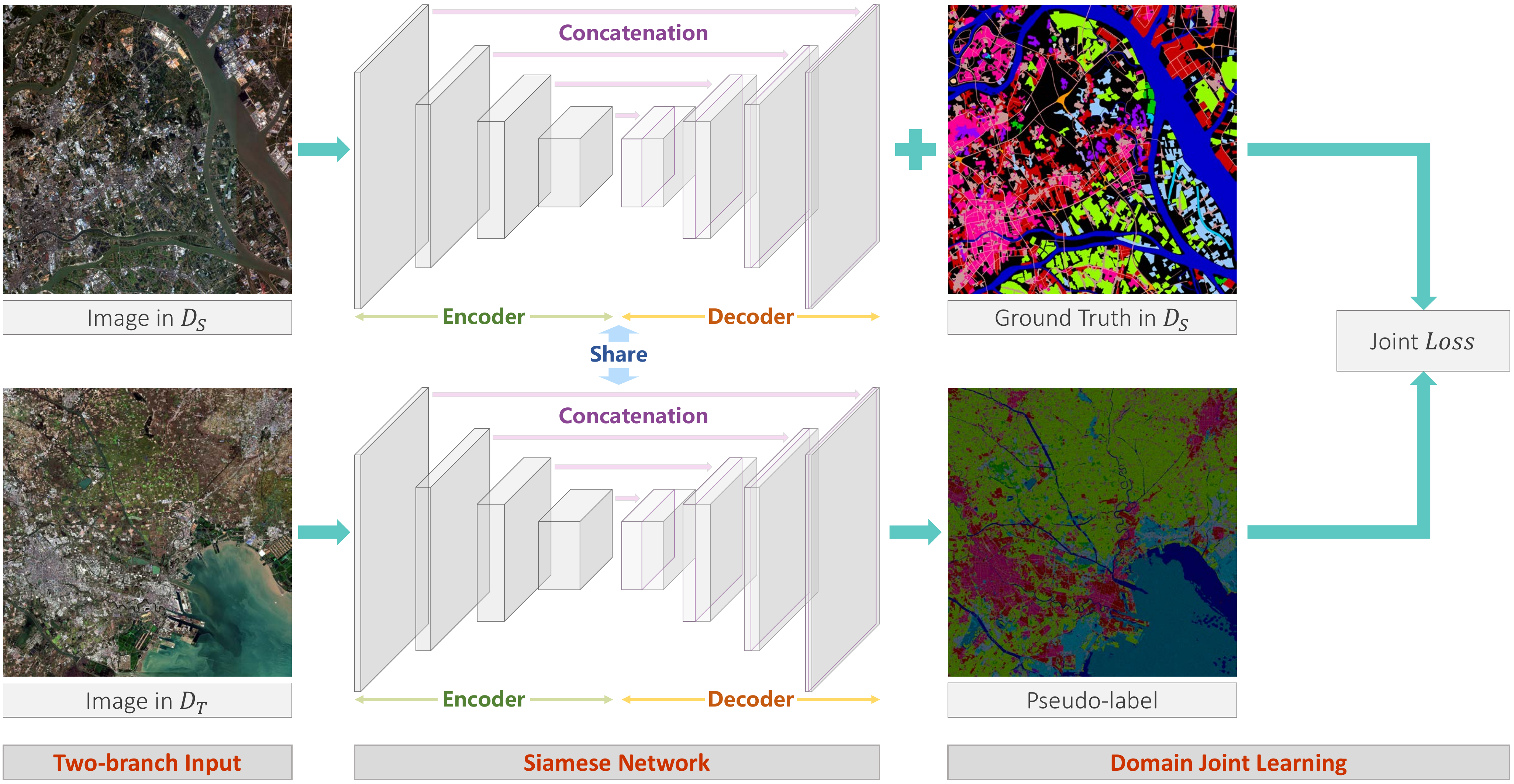}
\caption{Approach overview. We construct a Siamese network with two branches separately generating feature maps for images from $\textbf{\emph{D}}_{\textbf{\emph{S}}}$ and $\textbf{\emph{D}}_{\textbf{\emph{T}}}$. In the target domain branch, image pixels with high-confidence are assigned pseudo-labels. These pseudo-labels are then used to construct a joint classification loss with the source domain branch. U-Net is used as the backbone of the Siamese network.}
\label{figure:siameseunet}
\end{figure*}

\subsection{Semantic segmentation model for land cover classification}
\label{sec:segment}
There are two general strategies for deep-learning-based land cover classification: object-oriented approaches based on DCNNs and semantic segmentation approaches based on end-to-end DCNNs. The former ones use DCNNs to process images in the form of fixed-sized patches, and then distinguish deep features with shallow classifiers or directly employ the category predictions of DCNNs, to perform classification based on object-spatial units \cite{objectCNN1,objectCNN2,objectCNN3,GID}. In contrast, semantic segmentation models can predict dense classification maps for arbitrary-sized images in an end-to-end, pixels-to-pixels manner \cite{semanticFCN1,semanticFCN2,2019relation}.

The latest semantic segmentation models typically consist of two main paths: the encoder path that gradually reduces feature map size and captures higher-level information, and the decoder path that gradually recovers spatial resolution as well as clear object boundaries \cite{deeplab}. In our work, we adopt U-Net \cite{unet} as the backbone for land cover classification and domain adaptation. It is worth noting that U-Net specially achieves concatenation of each corresponding encoding and decoding stages using skip connection of feature maps, as illustrated in Fig. \ref{figure:siameseunet}, so that more raw information of the input image is retained and added to the decoding path. This design helps to compensate for the information loss in the encoding path, which is significant for the classification of satellite images that rely heavily on spectral information.

To enable U-Net to process MS remote sensing images, we adjust the channel number of its input to 4, i.e., we change the kernel size of its first convolutional layer from $3\times 3\times 3$ to $3\times 3\times 4$. In addition, we adjust the channel number of its output feature map according to our category system, i.e., we set the kernel number of its last convolutional layer to 24.

\subsection{Domain joint learning for unsupervised domain adaptation}
\label{sec:adaptat}
To adapt DCNNs to a new domain, there is no better way than having examples of its feature distribution \cite{DomainAdaptation}. Faced with $\textbf{\emph{D}}_{\textbf{\emph{T}}}$ without annotation information, we are inspired by pseudo-labeling \cite{pseudoLabel,GID} and propose a UDA approach that collects reliable pixel-wise examples from $\textbf{\emph{D}}_{\textbf{\emph{T}}}$ for model adaptation. Compared to discrepancy-based and adversarial-based UDA methods, which force the two distributions to be aligned in feature space, pseudo-labeling is more flexible and potentially more dependable for complicated real-world situations.

To prevent DCNNs from biasing toward incorrect pseudo-labels or categories of easy samples, our approach introduces a Siamese network (Section \ref{sec:methodnetwork}) to collect pseudo-labels of which the number is dynamically increased with training iterations (Section \ref{sec:methodpdlabel}). These pseudo-labels are used to train jointly with the true labels from $\textbf{\emph{D}}_{\textbf{\emph{S}}}$, and the joint classification loss is weighted according to the category distribution of $\textbf{\emph{D}}_{\textbf{\emph{S}}}$ (Section \ref{sec:methodcbjoint}).

\subsubsection{Siamese network}
\label{sec:methodnetwork}
To avoid introducing incorrect category information in the training, only a very small number of pseudo-labels are used in the initial iterations of domain joint learning, which leads to two problems: (1) $\textbf{\emph{D}}_{\textbf{\emph{T}}}$ can only provide very few training samples at the beginning; (2) the samples selected from $\textbf{\emph{D}}_{\textbf{\emph{T}}}$ may be extremely homogeneous. Therefore, to ensure that the parameters of DCNNs are effectively updated at each training iteration, we pre-train U-Net on $\textbf{\emph{D}}_{\textbf{\emph{S}}}$ and use it as the backbone to construct a Siamese network. Siamese network has two branches, each of which have an input and an output \cite{siamese}. The two branches have an identical architecture and share the same parameters during both initialization and training, which allows the Siamese network to learn information from two distributions simultaneously, as presented in Fig. \ref{figure:siameseunet}.

Formally, given $\textbf{\emph{D}}_{\textbf{\emph{S}}}\subset \mathbb{R}^{H\times W\times 4}$ along with associated labels $L_{\textbf{\emph{S}}}\subset [1,K]^{H\times W}$, and unlabeled $\textbf{\emph{D}}_{\textbf{\emph{T}}}\subset \mathbb{R}^{H\times W\times 4}$, where $H\times W$ indicates the size of images and label maps, $K$ is the total number of classes. The two branches of the Siamese network separately take images $x_{\textbf{\emph{S}}}\in \textbf{\emph{D}}_{\textbf{\emph{S}}}$ and $x_{\textbf{\emph{T}}}\in \textbf{\emph{D}}_{\textbf{\emph{T}}}$ and predict $K$-dimensional feature maps $F_{x_{\textbf{\emph{S}}}}\in \mathbb{R}^{H\times W\times K}$ and $F_{x_{\textbf{\emph{T}}}}\in \mathbb{R}^{H\times W\times K}$.

\begin{figure*}[htb!]
\centering
\includegraphics[width=0.7\textwidth]
{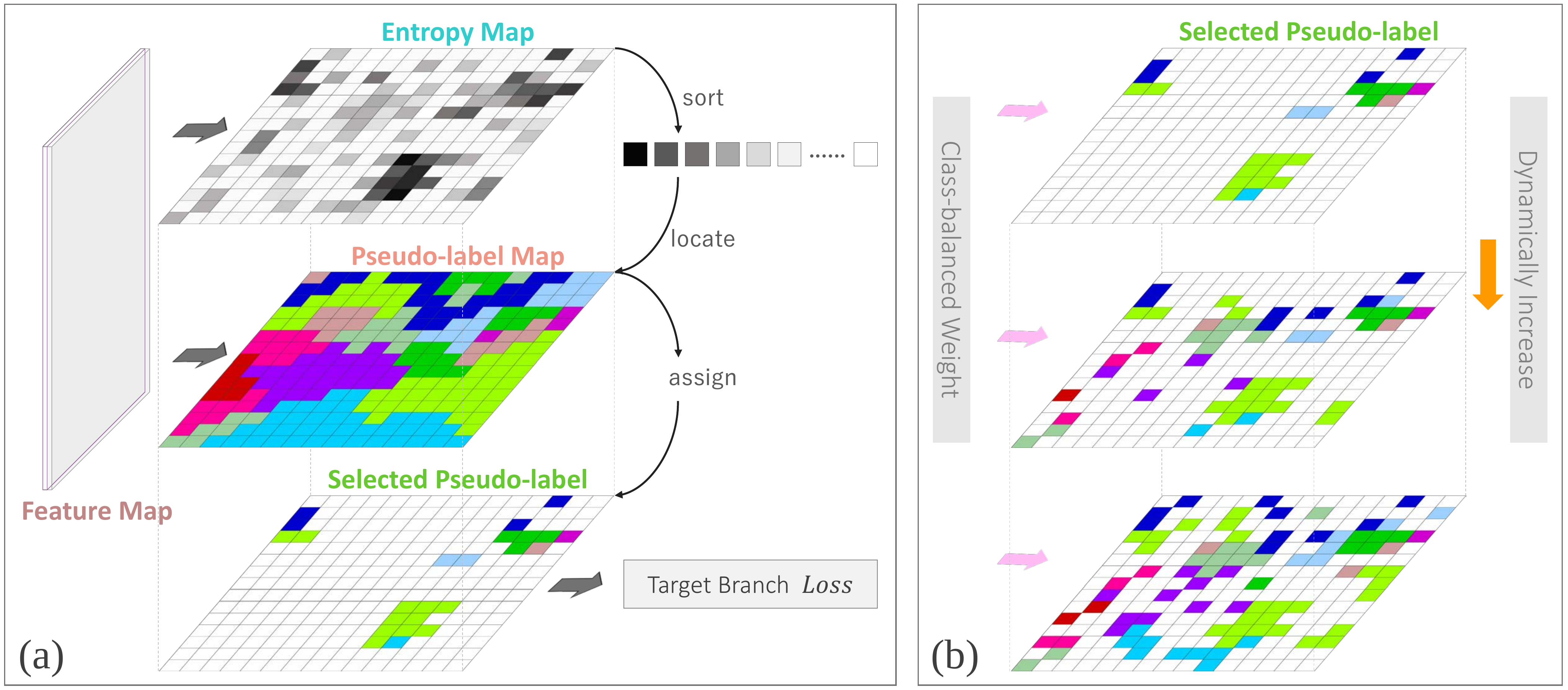}
\caption{(a) Pseudo-label assignment. (b) Dynamic labeling and class balancing. The number of pixels assigned pseudo-labels is dynamically changed with training iterations, and the joint classification loss is weighted according to the class distribution in $\textbf{\emph{D}}_{\textbf{\emph{S}}}$.}
\label{figure:pseudolabel}
\end{figure*}

\subsubsection{Dynamic pseudo-label assignment}
\label{sec:methodpdlabel}
Unlike the existing pseudo-labeling methods that empirically choose a threshold \cite{threshold1,threshold2} or set a fixed proportion for sample collection \cite{proportion1,proportion2,GID}, we assign pseudo-labels to a dynamic number of samples at different training epochs, as demonstrated in Fig. \ref{figure:pseudolabel}. Shannon Entropy \cite{threshold1} is employed as our indicator to quantify the confidence of each image pixel. Lower information entropy represents higher classification confidence. In the $\textbf{\emph{D}}_{\textbf{\emph{T}}}$ branch, entropy map $E_{x_{\textbf{\emph{T}}}}\in \mathbb{R}^{H\times W}$ is calculated as

\begin{equation}
E_{x_{\textbf{\emph{T}}}}^{(h,w)}=\frac{-1}{\log(K)}\sum_{k=1}^{K}F_{x_{\textbf{\emph{T}}}}^{(h,w,k)}\log(F_{x_{\textbf{\emph{T}}}}^{(h,w,k)}),
\end{equation}
where $E_{x_{\textbf{\emph{T}}}}^{(h,w)}\in [0,1]$ is the value of $E_{x_{\textbf{\emph{T}}}}$ at pixel $(h,w)$.

We arrange all the pixels in entropy map $E_{x_{\textbf{\emph{T}}}}$ in ascending order and select the first $N$ pixels, where

\begin{equation}
N=\lambda \cdot H\cdot W\frac{n_{e}}{N_{e}},
\end{equation}
where $n_{e}$ denotes that the current training is the $n_{e}$th epoch, $N_{e}$ is the total number of training epochs, and $\lambda$ is used to control the overall size of pseudo-labels. At the end of training, the proportion of selected pixels in all pixels of $x_{\textbf{\emph{T}}}$ is $\lambda$.

An intuitive interpretation of this design is that the network may give reliable predictions for only a small number of pixels when it is not adapted to $\textbf{\emph{D}}_{\textbf{\emph{T}}}$, and as the network gradually learns the distribution of $\textbf{\emph{D}}_{\textbf{\emph{T}}}$, it can make reliable predictions for an increasing number of pixels.

For the selected pixel located at $(h,w)$, the softmax function is used to obtain its category probability vector:

\begin{equation}
P_{x_{\textbf{\emph{T}}}}^{(h,w)}=\frac{\exp(F_{x_{\textbf{\emph{T}}}}^{(h,w)})}{\sum_{k=1}^{K}\exp(F_{x_{\textbf{\emph{T}}}}^{(h,w,k)})},
\end{equation}
where $F_{x_{\textbf{\emph{T}}}}^{(h,w)}\in \mathbb{R}^{K}$ is the feature vector of pixel located at $(h,w)$, and $P_{x_{\textbf{\emph{T}}}}^{(h,w)}\in \mathbb{R}^{K}$, of which the $k$th element represents the probability that this pixel belongs to class $k$.

And its pseudo-label is assigned as
\begin{equation}
l^{(h,w)}=\mathop{\arg\max}_{k\in\left\{1,\cdots,K\right\}}P_{x_{\textbf{\emph{T}}}}^{(h,w,k)},
\end{equation}
where $l^{(h,w)}\in\left\{1,\cdots,K\right\}$.

\subsubsection{Class-balanced domain joint training}
\label{sec:methodcbjoint}
Class balancing is a common strategy for the training of semantic segmentation models \cite{FCN}, but it is rarely used in UDA approaches because the category information in the target domain is unknown. Since we assign pseudo-labels to $\textbf{\emph{D}}_{\textbf{\emph{T}}}$, it is possible to reduce the distribution bias caused by unbalanced categories through this strategy.

For $\textbf{\emph{D}}_{\textbf{\emph{S}}}$, we count the ratio of the number of pixels in each category to the number of all labeled pixels. Supposing that the ratio of the class $k$ is $\mu _{k}$, its weight is

\begin{equation}
W_{k}=\frac{1}{\log(1+\mu_{k})}.
\end{equation}

Then, the loss function of the $\textbf{\emph{D}}_{\textbf{\emph{T}}}$ branch is calculated as

\begin{equation}
\mathcal{L}oss_{\textbf{\emph{D}}_{\textbf{\emph{T}}}}=\sum_{n=1}^{N}W_{l^{n}}\mathcal{F}_{CE}(l^{n},P_{x_{\textbf{\emph{T}}}}^{n}),
\end{equation}
where $\mathcal{F}_{CE}(\cdot )$ is the Cross Entropy loss function, $l^{n}$ and $P_{x_{\textbf{\emph{T}}}}^{n}$ denote the pseudo-label and the category probability vector of the $n$th pixel selected from $x_{\textbf{\emph{T}}}$, respectively.

If there are errors in pseudo-labels, a small number of mistakes may eventually lead to a relatively large bias during the iterative training. When gradually learning the distribution of $\textbf{\emph{D}}_{\textbf{\emph{T}}}$, to maintain the discrimination of the network for the true labels, we adopt joint learning of both the $\textbf{\emph{D}}_{\textbf{\emph{S}}}$ branch and the $\textbf{\emph{D}}_{\textbf{\emph{T}}}$ branch. The overall loss function of the Siamese network is

\begin{equation}
\mathcal{L}oss = \mathcal{L}oss_{\textbf{\emph{D}}_{\textbf{\emph{S}}}} + \mathcal{L}oss_{\textbf{\emph{D}}_{\textbf{\emph{T}}}},
\end{equation}
where $\mathcal{L}oss_{\textbf{\emph{D}}_{\textbf{\emph{S}}}}$ is calculated by all pixels of $x_{\textbf{\emph{S}}}$ and is also applied with class-balanced weighting.

When the training of the Siamese network is completed, forward propagation is performed on only one of the branches during the inference phase.

\section{Experiments}
Our experiments comprise two parts: (1) to explore the performance of different land cover classification approaches, we provide a benchmark on \emph{Five-Billion-Pixels} of three types of representative algorithms, including object-oriented classification based on spectral-spatial features, object-oriented classification based on deep learning, and semantic segmentation based on deep learning; (2) to validate the effectiveness of the proposed UDA approach, we perform practical land cover mapping on 11 cities using images from three different sensors. The implementation details, comparison approaches, and evaluation metrics are introduced in Section \ref{sec:exprsetup}. Section \ref{sec:benchmark} presents the benchmark on \emph{Five-Billion-Pixels}. Section \ref{sec:lcmapping} presents the results of land cover mapping.

\subsection{Experimental setup}
\label{sec:exprsetup}

\subsubsection{Setup for benchmark on Five-Billion-Pixels}
\label{sec:exprsetupbenchmark}
\textbf{Data Processing}: Since object-oriented approaches and semantic segmentation approaches have different requirements for training data, i.e., object-oriented approaches only allow each input sample to have one label, while semantic segmentation approaches require a label for each pixel of input sample, we prepare different training data for them. The \emph{Five-Billion-Pixels} dataset is randomly divided into a training set of 120 images and a test set of 30 images. For the two types of object-oriented approaches, we train the models using image patches with multiple scales \cite{GID}. The patch scales are set to the optimal values. Specifically, patches of sizes $64\times 64$ and $128\times 128$ pixels are randomly sampled from images of the training set. If more than $80\%$ pixels in a patch are covered by the same category, this patch is considered as a training sample. The ratio of the number of $64\times 64$-pixel patches to the number of $128\times 128$-pixel patches is $3:1$. In particular, for \emph{road}, the size of all patches is $32\times 32$ pixels. To balance the categories, we control the proportion of patches belonging to each class. A total of 130,000 multi-scale patches are randomly selected for model training. For semantic segmentation approaches, the original large images are cropped into image tiles with a size of $512\times 512$ pixels for model training. To improve the training efficiency, we only use tiles that are more than $50\%$ annotated and contain two or more categories. A total of 40,000 tiles are randomly selected from 120 training images.

\textbf{Baseline Methods}: For object-oriented classification based on spectral-spatial features, we employ multi-feature fusion strategy to aggregate spectral feature and gray-level co-occurrence matrix (GLCM) \cite{GLCM} by normalization and vector concatenation. Multi-layer perceptron (MLP) and random forest (RF) are utilized as classifiers. Selective search \cite{objectSegment} is adopted for object-spatial unit segmentation. The parameters of these methods are set to the optimal values. The window size of GLCM is $7\times 7$ pixels. MLP has 4 hidden layers with 20 nodes per layer. The number of trees for RF is 500. The initial segmentation size is 400 pixels for selective search. Classifiers are trained with image patches and are used to classify test images in units of objects.

For object-oriented classification based on deep learning, we employ two representative DCNNs: GoogLeNet \cite{GoogLeNet} and ResNet-101 \cite{resnet}. Both models are trained with the same hyper-parameters. The epoch number is 120, the batch size is 256, the momentum value is 0.9, and the weight decay is $10^{-4}$. The initial learning rate is 0.1 and is divided by 10 after every 30 epochs. In training, image patches are uniformly resized to $224\times 224$ pixels before being input to the models, and $20\%$ patches are used for model validation. Image augmentation strategies are adopted. In the testing phase, selective search is used for object segmentation with an initial segmentation size of 400 pixels. The test images are classified in units of $64\times 64$-pixel patches and then the patch-level classification map and the object-level segmentation map are combined via voting strategy \cite{GID}.

For semantic segmentation based on deep learning, we utilize U-Net \cite{unet} and DeepLabv3+ \cite{deeplab} as baseline models. The backbone chosen for DeepLabv3+ is ResNet-101 pre-trained on \emph{Five-Billion-Pixels}. And the out stride of DeepLabv3+ is set to 16. The weights of U-Net are randomly initialized. Both models are trained under the same conditions. The epoch number is 120, the batch size is 32, the momentum value is 0.9, the weight decay is $10^{-5}$. The initial learning rate is 0.05, and the poly learning policy \cite{poly} is used to adjust the learning rate during epochs. In the loss function, unlabeled regions are ignored, and class-balanced weighting is implemented according to Table \ref{table:ctgrratio}. During training, $20\%$ tiles are randomly selected for model validation. Image augmentation strategies are adopted. In the testing phase, models directly segment the test images in units of $512\times 512$-pixel tiles, and the overlap-tile strategy \cite{unet} is used to prevent context missing in the border region of tiles, where the overlap ratio is set to $50\%$.

\textbf{Evaluation Metrics}: We assess the experimental results with overall accuracy (OA), mean F1-score (mF1), mean intersection over union (mIOU), and user's accuracy (UA). mF1 is the category mean of F1-score. mIOU is the category mean of the intersection over union (IOU), and IOU is obtained by dividing the intersection of prediction and truth by their union \cite{FCN}. mF1 and mIOU describes the ability of the model to minimize both overestimation and underestimation for each category. UA indicates the performance of the model in reducing overestimation \cite{users}.

\subsubsection{Setup for land cover mapping}
\label{sec:exprsetuplcmapping}
\textbf{Data Processing}: To adapt deep models to satellite images with different resolutions, we construct a multi-scale source domain using \emph{Five-Billion-Pixels}. Image tiles with different sizes are randomly cropped from GF-2 according to the spatial resolution of multiple data sources, including $512\times 512$, $1024\times 1024$ (for GF-1), and $1280\times 1280$ (for ST-2) pixels, and are then uniformly resized to $512\times 512$ pixels. Since the $3$ m resolution of PS is obtained by resampling the raw data and its effective spatial resolution is $3.7$-$4.1$ m, we use the original image resolution of GF-2 to adapt it. The total number of tiles in the source domain is 12,800, and the ratio of the three sizes is $2:1:1$.

We prepare a data domain for each target city. The raw satellite images are cropped into non-overlapping tiles with a size of $512\times 512$ pixels. In particular, PS images are resized to $3/4$ of their original image resolution before cropping, equivalent to restoring their spatial resolution to $4$ m. The target domains of Beijing, Chengdu, Guangzhou, Shanghai, and Wuhan consist of 4126 ST-2 image tiles, 4144 PS image tiles, 1398 ST-2 image tiles, 3117 PS image tiles, and 1764 GF-1 image tiles, respectively. And the target domains of Bangkok, Delhi, Naypyidaw, Seoul, Tokyo, Japan, and Myanmar separately contains 441 ST-2 image tiles.

\textbf{Comparison Methods}: We compare our approach with the recent leading and representative UDA methods: AdaptSeg \cite{negativeAdapt}, AdvEnt \cite{Shannon}, CLAN \cite{CLAN}, and FADA \cite{FADA}, where AdaptSeg and CLAN are adversarial-based domain alignment methods, while AdvEnt and FADA combine adversarial domain alignment and pseudo-label learning. Specifically, AdaptSeg incorporates adversarial learning at different feature levels of the segmentation model; CLAN aligns each class with an adaptive adversarial loss to enforce local semantic consistency; AdvEnt minimizes the prediction entropy of the target domain using adversarial loss and pseudo-label loss; FADA implements fine-grained class-level feature alignment based on the class information of pseudo-labels. U-Net is used as the generator for these comparison methods. We also test the combination of these methods and our dynamic pseudo-label assignment approach by adding their loss functions. In addition, the baseline method is U-Net trained only with the source domain.

U-Net is initialized using the network parameters trained on \emph{Five-Billion-Pixels} (see Section \ref{sec:exprsetupbenchmark}). For our approach, the batch size is 16 for both the source and target branches (total 32). For the comparison approaches, the batch size is 32. This is because our approach inputs the source and target data simultaneously, while the comparison methods alternately inputs the source and target data. The initial learning rate for our approach is 0.001. And for the comparison approaches, the initial learning rate are 0.001 and 0.0001 for the generator and discriminator, respectively. For all methods, the epoch number is 100, the momentum is 0.9, the weight decay is $10^{-5}$, and the poly learning policy is used to adjust the learning rate during epochs. Image augmentation strategies are adopted. And class-balanced weighting is implemented according to Table \ref{table:ctgrratio}. For our approach, $\lambda$ (see Section \ref{sec:methodpdlabel}) is empirically set to 0.5.

To prevent different target domains from interfering with each other, we separately train a model for each city. Since the tile number of the source domain is much larger than those of the target domains, at each epoch in training, tiles of equal number to the target domain are randomly selected from the source domain. This results in a different sub-source domain at each epoch, allowing the model to select pixels with diversity from the target domain.

\subsection{Benchmark on Five-Billion-Pixels}
\label{sec:benchmark}
The baseline results for \emph{Five-Billion-Pixels} are listed in Table \ref{table:resultgid24}. It can be seen that deep-learning-based methods bring huge performance margins compared with methods based on spectral-spatial features and shallow classifiers. This shows that conventional methods lack discriminative ability for high-resolution images with complicated spatial information. 

\begin{table*}[htb!]
\caption{Benchmark on \emph{Five-Billion-Pixels}. The abbreviations for categories are defined as: Indu - \emph{industrial area}, Urba - \emph{urban residential}, Rura - \emph{rural residential}, Stad - \emph{stadium}, Squa - \emph{square}, Over - \emph{overpass}, Rail - \emph{railway station}, Airp - \emph{airport}, Padd - \emph{paddy field}, Irri - \emph{irrigated field}, Dryc - \emph{dry cropland}, Gard - \emph{garden land}, Arbo - \emph{arbor forest}, Shru - \emph{shrub forest}, Natu - \emph{natural meadow}, Arti - \emph{artificial meadow}, Rive - \emph{river}, Fish - \emph{fish pond}, Bare - \emph{bare land}. Accuracy results are expressed as percentage values ($\%$).}
\vspace{3mm}
\resizebox{\textwidth}{!}{
\begin{tabular}{lllllllllllllll} 
\hline
\textbf{Method}&\textbf{OA}&\textbf{mF1}&\textbf{\footnotesize{mIOU}}&\textbf{UA}:&Indu&Urba&Rura&Stad&Squa&Road&Over&Rail&Airp&Padd\\ 
\hline
MLP+Fusion & 23.89 & 15.81 &  9.78 && 48.49 & 38.58 & 13.69 &  0    &  0    &  9.27 &  0.76 &  0    &  0    & 22.90         \\
RF+Fusion  & 27.40 & 17.16 & 10.23 && 38.36 & 23.30 & 11.99 &  0.65 &  0.41 & 10.39 &  0.80 &  0.05 &  3.13 & 22.21         \\
GoogLeNet  & 69.19 & 39.70 & 28.99 && 51.07 & 66.68 & 71.95 & 78.30 &  8.54 & 37.87 & 14.35 & 15.70 & 34.92 & 47.58         \\
ResNet101  & 69.55 & 45.73 & 33.59 && 58.44 & 69.22 & 70.89 & 82.55 &  8.93 & 42.70 & 12.20 & 27.32 &\textbf{54.05}&50.71   \\
DeepLabv3+ & 79.87 & 54.84 & 42.12 && 76.87 & 74.89 & 79.80 &\textbf{86.36} &\textbf{18.11} & 82.59 &\textbf{58.50}&\textbf{56.16}&24.40&64.53\\
U-Net      &\textbf{80.35}&\textbf{57.34}&\textbf{44.51}&&\textbf{80.72}&\textbf{83.88}&\textbf{85.73}&47.32&15.68&\textbf{84.15}&43.68&41.28&34.43&\textbf{74.34}\\ 
\hline
           &Irri&Dryc&Gard&Arbo&Shru&Park&Natu&Arti&Rive&Lake&Pond&Fish&Snow&Bare\\ 
\hline
MLP+Fusion & 47.11 & 20.85 &  6.44  & 54.01 &  0.58 &  0    & 33.21 &  2.51 & 52.99 & 73.68 &  5.91 & 22.35 &  0.13 &  0     \\
RF+Fusion  & 59.05 & 29.21 &  2.61  & 58.53 &  0.43 &  0    & 37.22 &  3.65 & 44.96 & 65.04 &  7.88 & 31.98 &  4.50 & 21.47  \\
GoogLeNet  & 85.25 & 77.70 & 10.29  & 82.88 & 12.89 & 18.14 & 64.47 & 51.88 & 72.53 & 71.60 &  9.89 & 55.53 &  0    & 71.78  \\
ResNet101  & 87.45 & 79.20 & 14.23  & 86.61 & 17.08 & 18.40 & 69.77 & 64.59 & 67.01 & 69.28 &  9.45 & 59.41 & 52.44 & 78.47  \\
DeepLabv3+ & 87.80 & 79.18 & 14.40  & 94.38 & 19.80 &\textbf{56.49} & 81.75 &\textbf{81.00} &\textbf{91.71} &\textbf{75.12}&21.23&76.13&\textbf{86.60}&90.57\\
U-Net      &\textbf{88.86}&\textbf{81.30}&\textbf{38.05}&\textbf{95.42}&\textbf{25.39}&42.92&\textbf{87.10}&68.66&63.58&70.48&\textbf{21.89}&\textbf{78.75}&35.83&\textbf{96.17}\\
\hline
\end{tabular}}
\label{table:resultgid24}
\end{table*}

Within deep-learning-based methods, semantic segmentation models (U-Net and DeepLabv3+) significantly outperform object-oriented methods (ResNet101 and GoogLeNet). This is because semantic segmentation models can capture contextual information over larger areas and simultaneously maintain more accurate edges for ground objects by assigning labels to each pixel. The performance advantages of deep learning, especially of semantic segmentation models, demonstrate the importance of large-scale, pixel-wise annotated datasets for advancing land cover classification research.

ResNet101 behaves better than GoogLeNet in overall and has significantly superior results on \emph{railway station}, \emph{airport}, \emph{shrub forest}, \emph{natural meadow}, and \emph{artificial meadow}. Owing to the residual connection structure \cite{resnet} that enables the combination of different levels of features, ResNet101 can learn low-level features to distinguish natural classes as well as high-level features to identify artificial buildings with complex structures.

DeepLabv3+ achieves the best results for urban functional areas, including \emph{stadium}, \emph{square}, \emph{overpass}, \emph{railway station}, \emph{park}, and \emph{artificial meadow}, which contain complicated spatial structures. Due to the atrous convolution \cite{atrous} and spatial pyramid pooling \cite{deeplab} adopted in DeepLabv3+, it can capture multi-level contextual information for these categories. U-Net behaves best on \emph{industrial area}, \emph{urban residential}, \emph{rural residential}, different agriculture, and different forest classes. The recognition of these categories relies heavily on textural and spectral information. U-Net has stronger discriminative ability for them because it retains more raw image information through the concatenation structure \cite{unet}.

Another issue worth noting is that regardless of the method type, there are high performance discrepancies of different categories. For instance, all methods behave poorly on \emph{square}, \emph{overpass}, \emph{railway station}, \emph{airport}, \emph{garden land}, \emph{park}, and \emph{pond}. This is due to two factors, first, these categories represent small percentages in the \emph{Five-Billion-Pixels} dataset, and second, they are inherently easier to confuse with other categories. These classes cover a much smaller area in the cities compared to residential and agricultural categories, and the models will be biased towards common and simple classes in training. In addition, the distinctive characteristic of these categories is that they are composed of multiple basic ground cover types; for example, \emph{railway station} contains multiple tracks and stadium-like building roof, \emph{park} includes grass and woods, and \emph{airport} contains roads and lawns, which causes them to be easily misclassified into other categories. This is why even if we control the proportion of different categories to be consistent in the object-oriented approachs (see Section \ref{sec:exprsetupbenchmark}), their accuracy still cannot be improved.

\begin{figure*}[htb!]
\centering
\includegraphics[width=0.7\textwidth]
{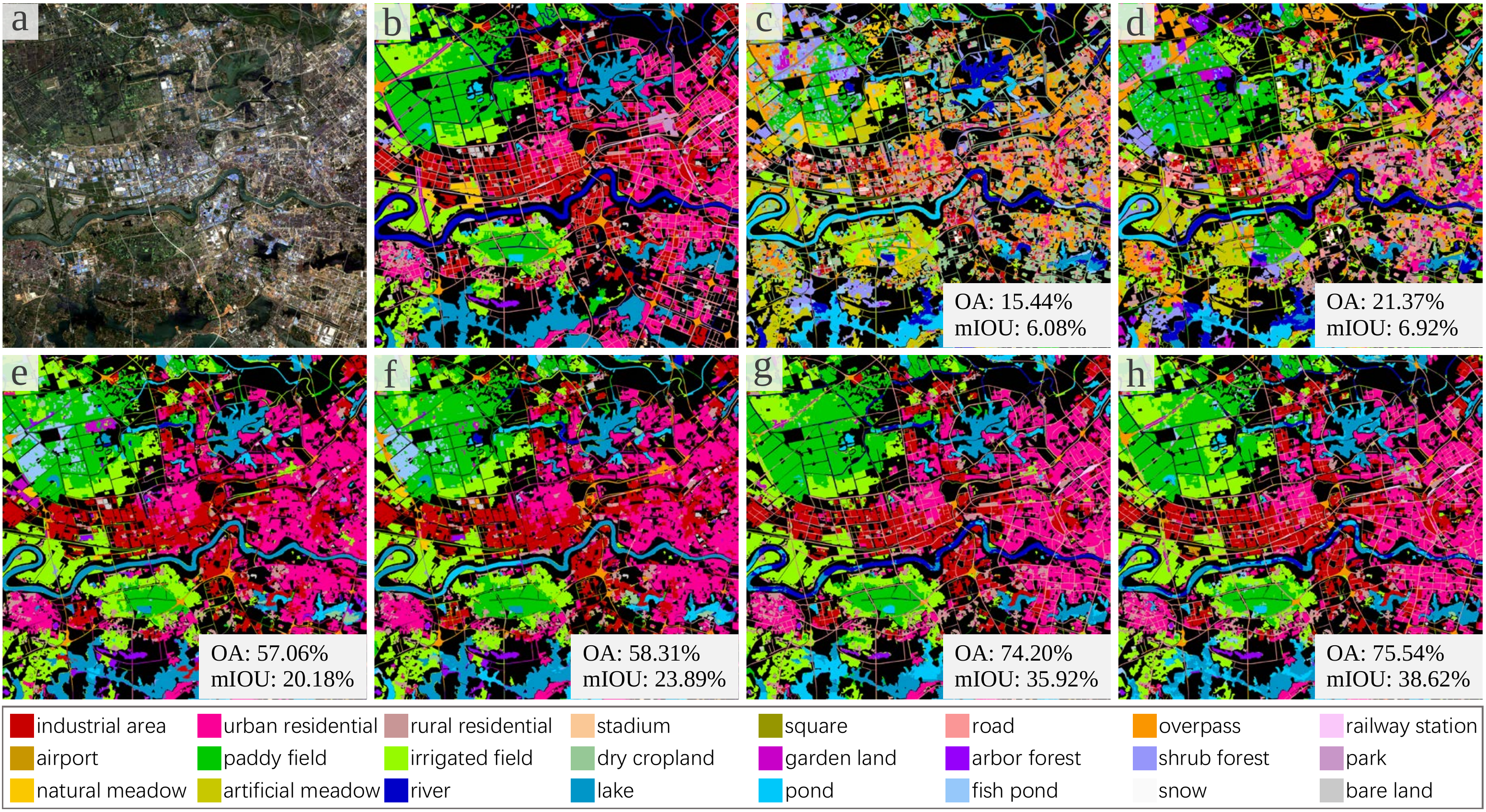}
\caption{A set of land cover classification maps of different baseline methods. (a) Input GF-2 satellite image. (b) Ground truth, where black indicates unlabeled areas. (c)–(h) Results of MLP+Fusion, RF+Fusion, GoogLeNet, ResNet101, DeepLabv3+, and U-Net, respectively.}
\label{figure:resultgid24}
\end{figure*}

To demonstrate the results more intuitively, a set of land cover classification maps is displayed in Fig. \ref{figure:resultgid24}. MLP+Fusion can identify some water areas, RF+Fusion can identify some water, built-up, and \emph{paddy field} areas, but the rest of the map is heavily confused. GoogLeNet and ResNet101 fail in extracting \emph{road} and misclassify \emph{paddy field} into \emph{fish pond}. In contrast, DeepLabv3+ and U-Net can segment clear \emph{road} networks and different built-up areas. DeepLabv3+ performs better on \emph{river} than U-Net. And U-Net can recognize \emph{lake} and \emph{irrigated field} more accurately.

\subsection{Land cover mapping}
\label{sec:lcmapping}

\subsubsection{Experimental results of Chinese megacities}

\begin{figure*}[htb!]
\centering
\includegraphics[width=0.7\textwidth]
{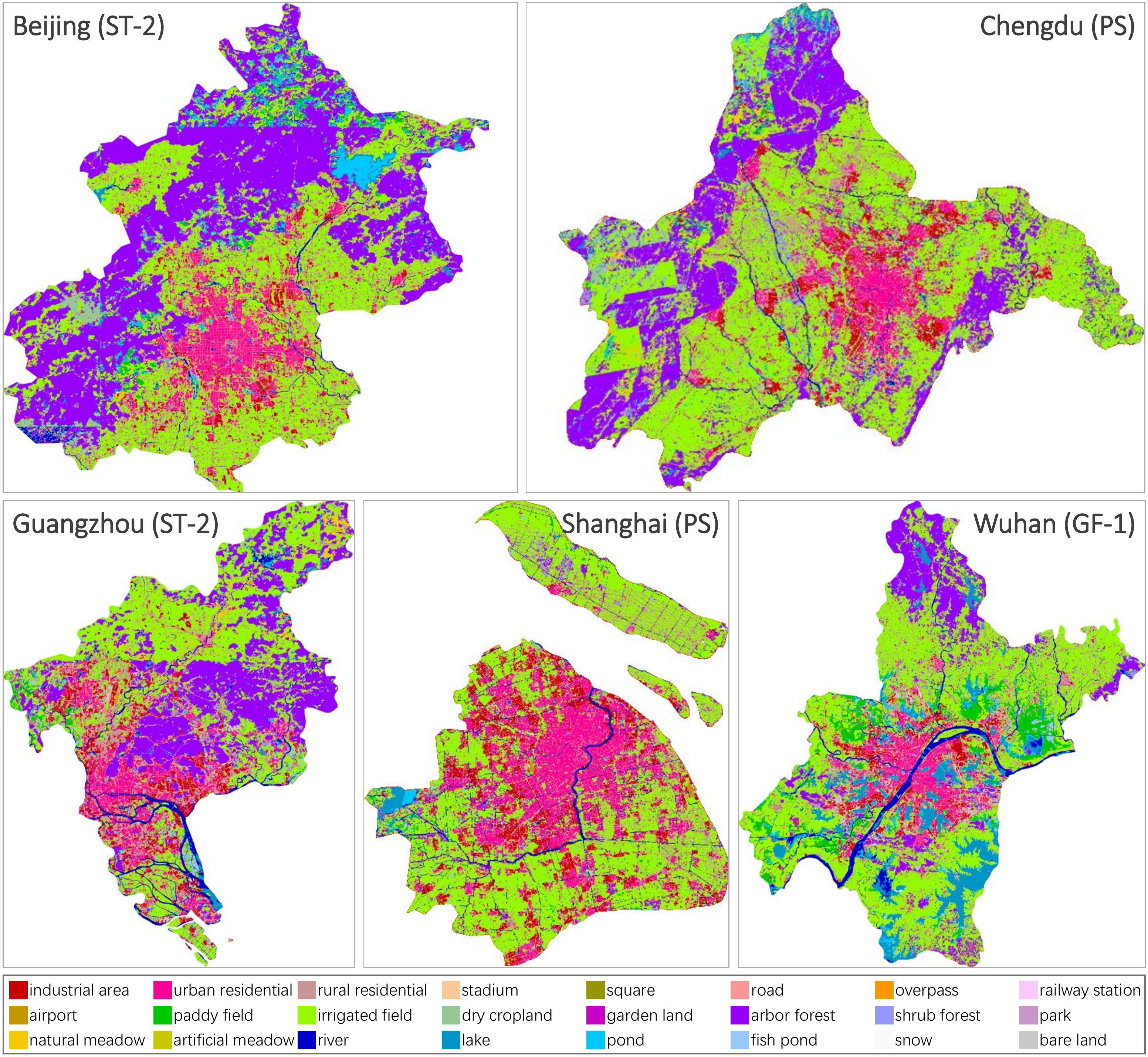}
\caption{Land cover mapping results for five Chinese megacities. Clear, detailed resulting maps can be found at \url{https://x-ytong.github.io/project/Five-Billion-Pixels.html}.}
\label{figure:resultlarge}
\end{figure*}

The land cover maps of five Chinese megacities are demonstrated in Fig. \ref{figure:resultlarge}. Although images from different sensors are utilized, and there is no annotated information on them, our approach is able to distinguish \emph{industrial area}, \emph{urban residential} in city center, \emph{rural residential} dispersed in suburb, transportation networks, and \emph{river} systems for each city. \emph{Fish pond} along the coast of Guangzhou and \emph{paddy field} in the suburbs of Wuhan are correctly identified. Obvious errors occur in the forested mountains and at the mosaic borders. Small areas of mountains around Beijing are misclassified as water bodies, and some areas of mountains around Chengdu are misclassified as \emph{irrigated field}. This is caused by the severe spectral shifts due to different image sources and different imaging conditions. Unlike artificial-constructed categories, the classification of natural classes relies more on spectral information. When the spectral shifts are particularly significant, pseudo-labels inevitably contain errors, which will continuously accumulate in iterative domain adaptation learning.

\begin{figure*}[htb!]
\centering
\includegraphics[width=0.82\textwidth]
{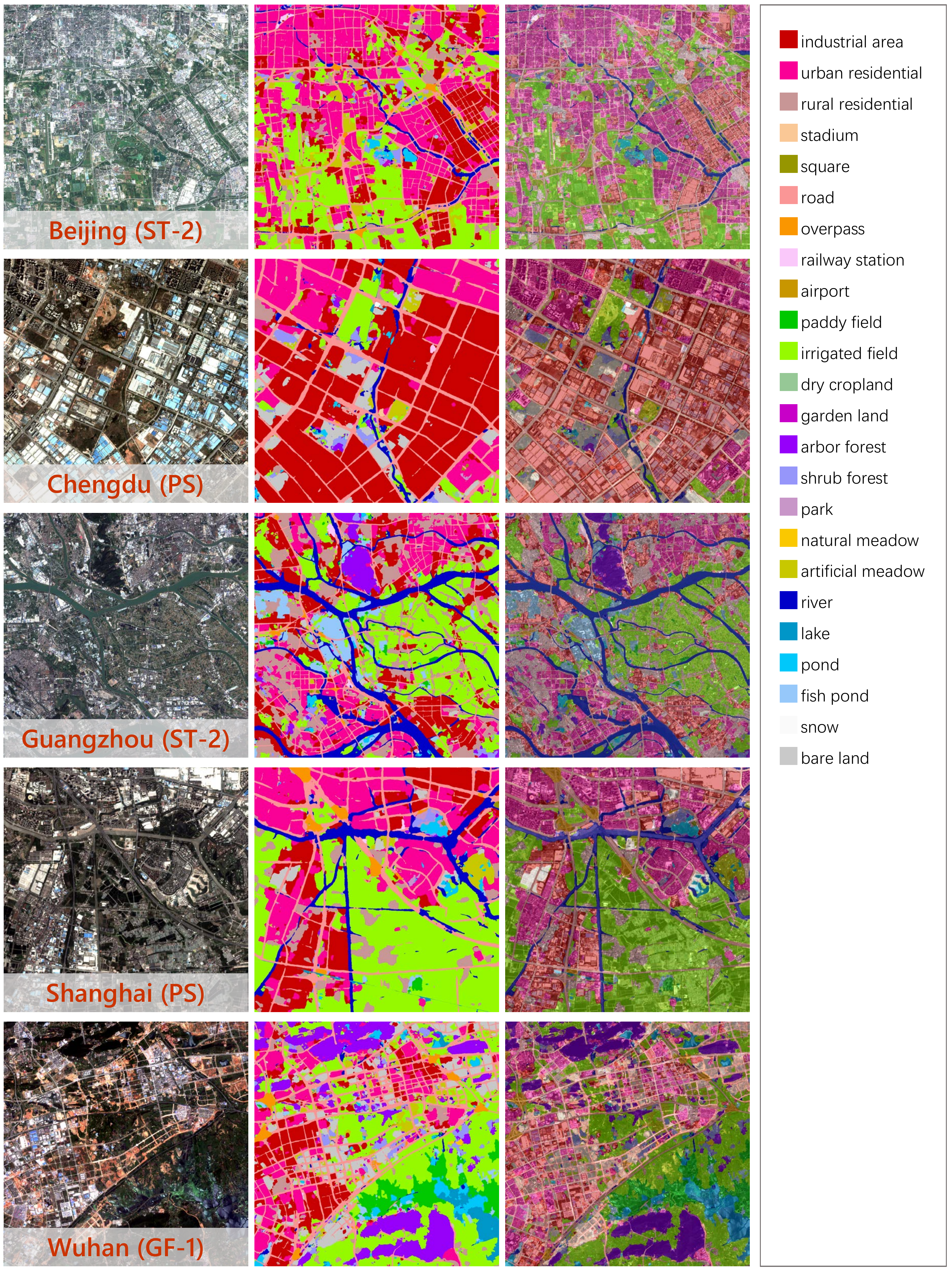}
\caption{Details of mapping results of different Chinese megacities. The first column shows partial regions of the input satellite images; the second column shows the corresponding classification results. And in the third column, the classification results are overlaid on the input images, which demonstrates the performance more visually.}
\label{figure:resultmulti}
\end{figure*}

Table \ref{table: resultmegacity} displays the quantitative evaluation results based on different test strategies. It can be seen that the results on dense label are generally better than that on sparse label in OA, while sparse label outperforms dense label in mF1 and mIOU. This is because mF1 and mIOU are more sensitive to overestimation and underestimation, in other words, the edges of the ground objects. And sparse label marks only portions of the ground objects, while dense label strictly outlines the edges of the ground objects, which leads to poorer mF1 and mIOU results on dense label. Whereas OA is the accuracy of the entire test area, and sparse label which is distributed evenly over the entire images contains more areas that are difficult to identify, e.g. complex urban functional areas, it therefore has lower OA values. More classification details are displayed in Fig. \ref{figure:resultmulti}. Our approach achieves promising performance on different built-up, traffic, and agricultural classes, as well as \emph{river}, and \emph{bare land}.

\begin{table*}[htb!]
\centering
\caption{Quantitative evaluation of land cover mapping for five Chinese megacities based on sparse label and dense label. Accuracy results are expressed as percentage values ($\%$).}
\vspace{3mm}
\resizebox{0.53\textwidth}{!}{
\begin{tabular}{lllllllll} 
\hline
\textbf{Megacity} &  &\multicolumn{3}{l}{\textbf{Sparse Label}}&&\multicolumn{3}{l}{\textbf{Dense Label}}\\
                    &  & OA & mF1 & \small{mIOU} &  & OA & mF1 & \small{mIOU}  \\ 
\hline
Beijing             &  & 70.86 & 49.76 & 39.72 &  & 87.45 & 42.43 & 33.02 \\
Chengdu             &  & 71.21 & 51.06 & 39.70 &  & 76.64 & 35.99 & 26.03 \\
Guangzhou           &  & 71.23 & 48.56 & 39.52 &  & 81.97 & 46.25 & 38.09 \\
Shanghai            &  & 74.80 & 56.84 & 44.19 &  & 74.70 & 48.22 & 39.50 \\
Wuhan               &  & 82.29 & 63.12 & 52.59 &  & 85.62 & 61.06 & 50.14 \\
\hline
\end{tabular}}
\label{table: resultmegacity}
\end{table*}

Table \ref{table:resultmulti} shows the quantitative evaluation of different UDA approaches, where all accuracy results are averaged across the five megacities. It can be seen that $\textbf{\emph{D}}_{\textbf{\emph{S}}}$ constructed by multi-scale GF-2 image tiles brings a significant improvement compared to U-Net pre-trained with single-scale tiles. This indicates that it is feasible to adapt models to images with different spatial resolutions by using multi-scale source domain data. There is a decline in performance of AdaptSeg, AdvEnt, and CLAN compared to the baseline ($\textbf{\emph{D}}_{\textbf{\emph{S}}}$-only). This is due to the negative adaptation caused by the complex feature distributions of both the source and target domains. In addition, the size of $\textbf{\emph{D}}_{\textbf{\emph{S}}}$ is much larger than that of $\textbf{\emph{D}}_{\textbf{\emph{T}}}$, and each training epoch uses a different sub-source domain (see Section \ref{sec:exprsetuplcmapping}), bringing more confusion into UDA approaches based on domain distribution alignment. FADA performs better than other comparison methods because FADA aligns intermediate level features rather than deep pixel-level features, which avoids rigid global matching. In addition, FADA achieves fine-grained class-level feature alignment according to the category information of pseudo-labels, thus alleviating the negative adaptation caused by category imbalance. The combination of these methods and pseudo-label assignment improves the accuracy, but is inferior to our method in all evaluation results, indicating that pseudo-labeling is not enough to compensate for the negative adaptation caused by domain distribution alignment.

\begin{table*}[htb!]
\centering
\caption{Comparison with the recent leading UDA approaches on Chinese megacities. ``Pre-trained'' means U-Net pre-trained with single-scale GF-2 image tiles, and PS images keep the original scale for this strategy. ``$\textbf{\emph{D}}_{\textbf{\emph{S}}}$-only'' is the baseline, meaning only using the multi-scale $\textbf{\emph{D}}_{\textbf{\emph{S}}}$ to train the Siamese network. ``DPA'' indicates dynamic pseudo-label assignment. Accuracy results are averaged over the five megacities and expressed as percentage values ($\%$).}
\vspace{3mm}
\resizebox{0.6\textwidth}{!}{
\begin{tabular}{lllllllll} 
\hline
\textbf{Method} &  &\multicolumn{3}{l}{\textbf{Sparse Label}} &&\multicolumn{3}{l}{\textbf{Dense Label}}\\
                &  & OA & mF1 & \small{mIOU} && OA & mF1 & \small{mIOU} \\ 
\hline
Pre-trained                                 && 70.56 & 46.33 & 37.64 && 75.42 & 38.93 & 28.25\\
$\textbf{\emph{D}}_{\textbf{\emph{S}}}$-only&& 72.65 & 49.95 & 40.01 && 78.19 & 42.31 & 32.55\\
AdaptSeg                                    && 67.18 & 42.28 & 31.18 && 73.00 & 36.96 & 26.82\\
AdaptSeg+DPA                                && 70.91 & 46.90 & 33.92 && 76.80 & 39.18 & 28.90\\
AdvEnt                                      && 64.51 & 41.50 & 30.49 && 75.18 & 35.29 & 26.49\\
AdvEnt+DPA                                  && 68.69 & 45.42 & 32.51 && 76.85 & 38.58 & 28.62\\
CLAN                                        && 65.05 & 42.18 & 31.07 && 72.92 & 35.71 & 26.74\\
CLAN+DPA                                    && 69.68 & 45.65 & 32.84 && 75.99 & 38.57 & 28.49\\
FADA                                        && 69.83 & 50.86 & 39.57 && 78.64 & 41.85 & 33.68\\
FADA+DPA                                    && 73.95 & 53.15 & 41.03 && 81.23 & 45.26 & 36.33\\
Ours (DPA) &&\textbf{74.08}&\textbf{53.87}&\textbf{43.14}&&\textbf{81.28}&\textbf{46.79}&\textbf{37.36}\\
\hline
\end{tabular}}
\label{table:resultmulti}
\end{table*}

\begin{figure*}[htb!]
\centering
\includegraphics[width=0.7\textwidth]
{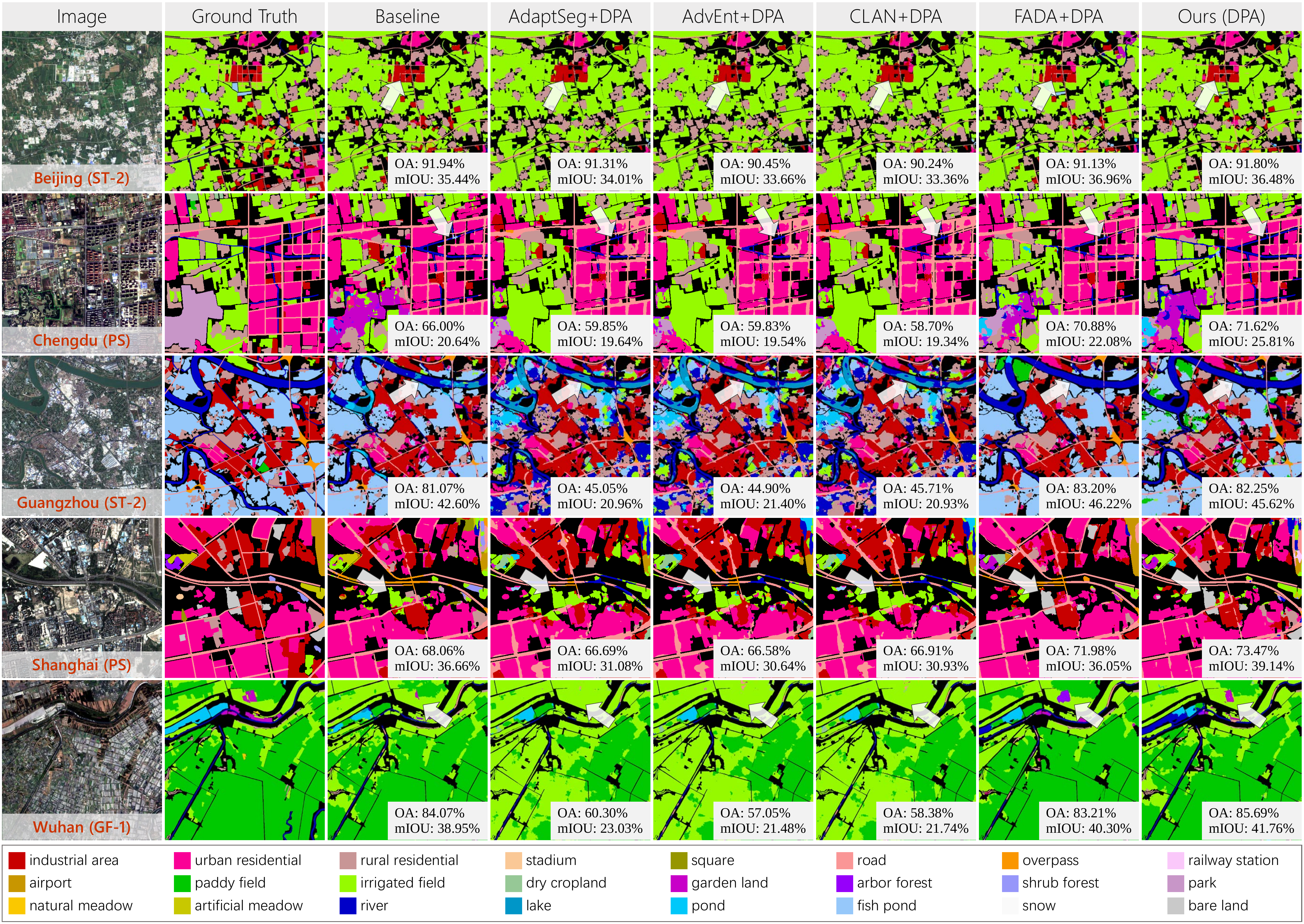}
\caption{Details of performance of different UDA approaches on Chinese megacities. ``Baseline'' means only using $\textbf{\emph{D}}_{\textbf{\emph{S}}}$ to train the Siamese network. ``DPA'' indicates dynamic pseudo-label assignment.}
\label{figure:resultadapt}
\end{figure*}

Fig. \ref{figure:resultadapt} illustrates details of the results obtained by different UDA methods. Here we demonstrate the performance of combination of AdaptSeg, AdvEnt, CLAN, and FADA with pseudo-labeling, which behaves better than these methods alone. In Beijing and Chengdu, the comparison methods lost fine \emph{road} and \emph{river} in built-up areas. And in Guangzhou, the comparison methods misclassify \emph{fish pond} as \emph{river}, \emph{pond}, and \emph{irrigated field}. This is because these misclassified categories are much less common, and the adversarial-based methods match domain distributions in a global manner, causing the aligned feature space to tend to prioritize the categories with larger sample amounts. In Shanghai and Wuhan, the results of \emph{bare land}, \emph{road}, \emph{gerden land}, and \emph{paddy field} are significantly improved by our approach compared with the baseline. This indicates that our approach can learn information of the target domain while maintaining the ability to identify the distribution of both domains. In Chengdu, our approach misclassifies \emph{park} into \emph{pond} and \emph{gerden land}, which are the ground objects contained by the park. It is probably because pseudo-labels in pixels are difficult to capture contextual information, and the adapted model is more biased towards the categories in local regions. In addition, it can be seen that for different cities, the improvements of our approach compared to the baseline are different, which is caused by the differences in feature distributions and category distributions. When the features in the target and source domains are quite distinct, the pseudo-labels are more ``valuable'' in domain joint training and can improve the transferability of the model to the target domain more significantly. And when the category distribution of the test area is very unbalanced, pseudo-labels of the hard samples can also lead to greater performance improvements.

\begin{figure*}[htb!]
\centering
\includegraphics[width=0.7\textwidth]
{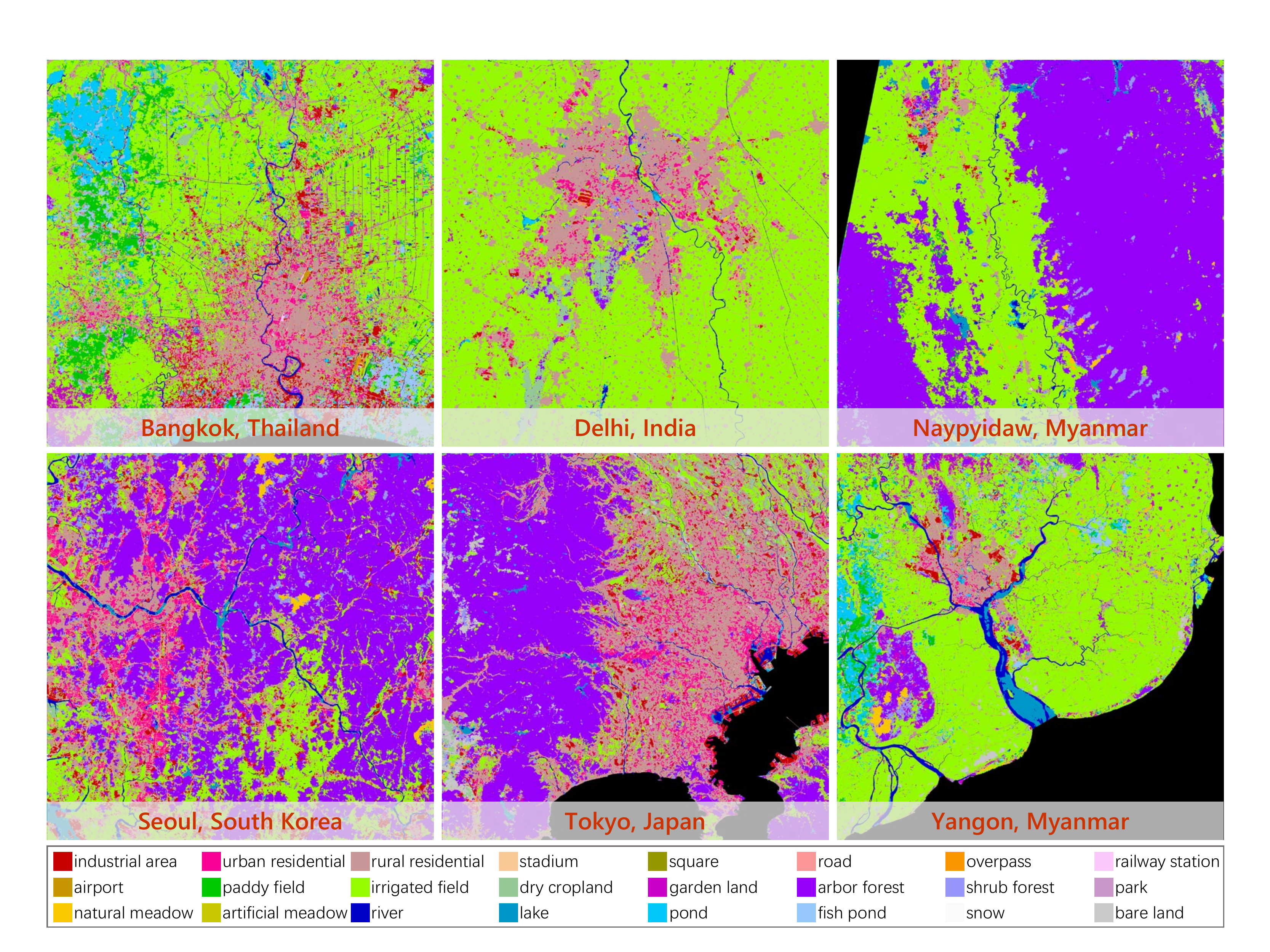}
\caption{Land cover mapping results for the additional six Asian cities.}
\label{figure:resultasia}
\end{figure*}

\subsubsection{Experimental results of additional Asian cities}
Fig. \ref{figure:resultasia} shows the results of land cover mapping for the additional six Asian cities. In particular, ``rural residential'' here refers to low-rise residential buildings, and ``urban residential'' refers to high-rise residential buildings. Although the target and source domains of this set of experiments are located in different countries with different geographical environment and urban landscapes, promising results are achieved. \emph{Paddy field} and \emph{garden land} located in the suburbs of Bangkok and Yangon are correctly identified, \emph{rural residential} areas spread around Delhi are accurately extracted, and the dense built-up areas of Seoul and Tokyo are well distinguished. Obvious errors occur in water bodies, where \emph{lake} and \emph{river} are heavily confused. Part of \emph{arbor forest} in Naypyidaw is misclassified as \emph{irrigated field}. In addition, the extraction results of \emph{road} are much less refined than those in Chinese cities.

\begin{table*}[htb!]
\centering
\caption{Comparison with the recent leading UDA approaches on the additional six Asian cities. ``$\textbf{\emph{D}}_{\textbf{\emph{S}}}$-only'' is the baseline, meaning only using the multi-scale $\textbf{\emph{D}}_{\textbf{\emph{S}}}$ to train the Siamese network. ``DPA'' indicates dynamic pseudo-label assignment. Accuracy results are averaged over the six cities and expressed as percentage values ($\%$).}
\vspace{3mm}
\resizebox{0.6\textwidth}{!}{
\begin{tabular}{lllllllll} 
\hline
\textbf{Method} &  &\multicolumn{3}{l}{\textbf{Sparse Label}} &&\multicolumn{3}{l}{\textbf{Dense Label}}\\
                &  & OA & mF1 & \small{mIOU} && OA & mF1 & \small{mIOU} \\ 
\hline
$\textbf{\emph{D}}_{\textbf{\emph{S}}}$-only&& 79.59 & 48.35 & 37.81 && 76.23 & 50.98 & 38.75 \\
AdaptSeg                                    && 61.19 & 26.02 & 18.23 && 58.58 & 30.23 & 20.61 \\
AdaptSeg+DPA                                && 64.46 & 32.35 & 19.21 && 60.80 & 34.94 & 22.37 \\
AdvEnt                                      && 61.30 & 26.61 & 17.64 && 60.35 & 31.37 & 21.58 \\
AdvEnt+DPA                                  && 63.94 & 30.73 & 19.58 && 62.23 & 35.28 & 22.85 \\
CLAN                                        && 62.70 & 25.85 & 17.91 && 58.98 & 30.33 & 20.77 \\
CLAN+DPA                                    && 64.66 & 31.27 & 19.20 && 61.08 & 34.35 & 22.38 \\
FADA                                        && 68.43 & 42.39 & 32.95 && 77.62 & 40.46 & 32.81 \\
FADA+DPA                                    && 70.86 & 46.84 & 35.74 && 79.21 & 43.50 & 34.35 \\
Ours (DPA) &&\textbf{81.14}&\textbf{51.20}&\textbf{40.81}&&\textbf{80.85}&\textbf{55.33}&\textbf{43.99}\\
\hline
\end{tabular}}
\label{table:resultasia}
\end{table*}

\begin{figure*}[htb!]
\centering
\includegraphics[width=0.7\textwidth]
{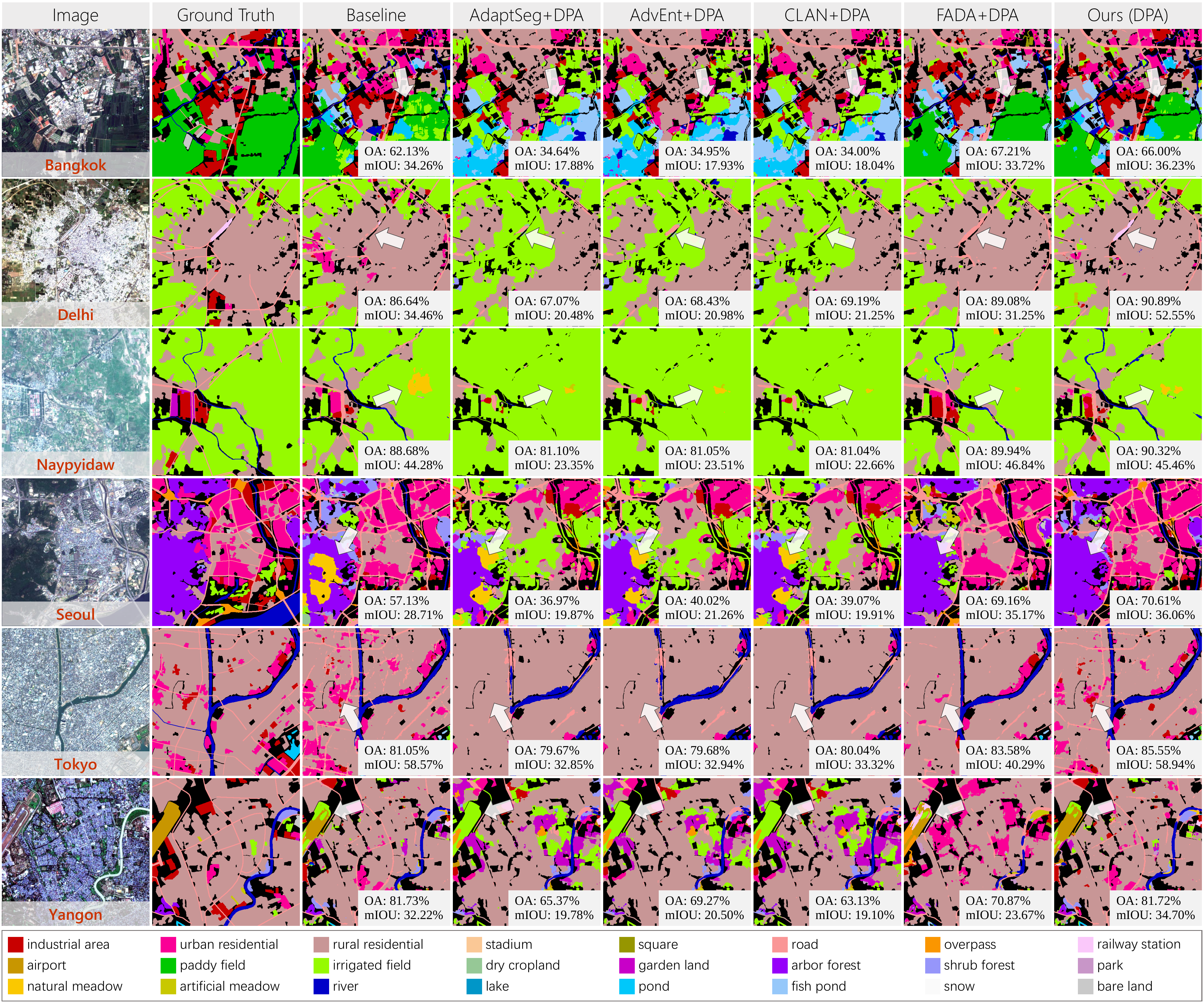}
\caption{Details of results of different UDA approaches on the additional Asian cities. ``Baseline'' means only using $\textbf{\emph{D}}_{\textbf{\emph{S}}}$ to train the Siamese network. ``DPA'' indicates dynamic pseudo-label assignment.}
\label{figure:resultsub}
\end{figure*}

The quantitative evaluation of land cover mapping are presented in Table \ref{table:resultasia}, where all accuracy results are averaged across the six cities. Compared to the baseline, our approach obviously boosts the performance, which indicates that our approach can mine reliable pseudo-labels even for very different geographical regions. The performance gap between the comparison methods and our approach is greater in these six cities than in Chinese megacities. This is because in this scenario, the feature distributions of the source and target domains are more disparate, it is harder for the adversarial-based approaches to find a suitable match between the two distributions, and more intra-domain variance is introduced into the model by the rigid alignment. Another phenomenon is that the results on sparse label and dense label are generally comparable. There are two reasons for this phenomenon, on the one hand, the spatial resolution of ST-2 is lower than those of PS and GF-1, and it cannot present very fine edges in both ground truth and results. On the other hand, agricultural and natural categories that are easier to classify occupy a larger total area in these six cities, and there are fewer complex urban functional classes located in the test areas. Therefore, neither sparse label nor dense label can pull apart the gap between overall accuracy and edge accuracy.

More details can be seen in Fig. \ref{figure:resultsub}, our approach improves the results of \emph{paddy field} in Bangkok, \emph{railway station} in Delhi, \emph{irrigated field} in Naypyidaw, \emph{arbor forest} in Seoul, \emph{industrial area} in Tokyo, and \emph{airport} in Yangon. However, even though the overall accuracy is encouraging, \emph{road} lines in this set of results are discontinuous. This is due to the feature distribution differences caused by the diverse architectural styles and city landscapes in different countries. The adversarial-based UDA methods demonstrate severe negative adaptation; they tend to classify complex areas into common categories when the distributions of the source and target domains are very different. For instance, in Delhi, Naypyidaw, and Seoul, the comparison methods classify built-up areas into \emph{irrigated field}, and in Tokyo, \emph{high-rise residential} and \emph{industrial area} are misclassified as \emph{low-rise residential}.

\subsubsection{Sensitivity analysis}
Some parameters have an impact on the domain adaptation results; we analyze these factors on the sparse label of five Chinese megacities in this section, including the number of training epochs and $\lambda$ (see Section \ref{sec:methodpdlabel}).

\begin{figure*}[htb!]
\centering
\subfigure[Epochs]
{\includegraphics[width=0.3\textwidth]
{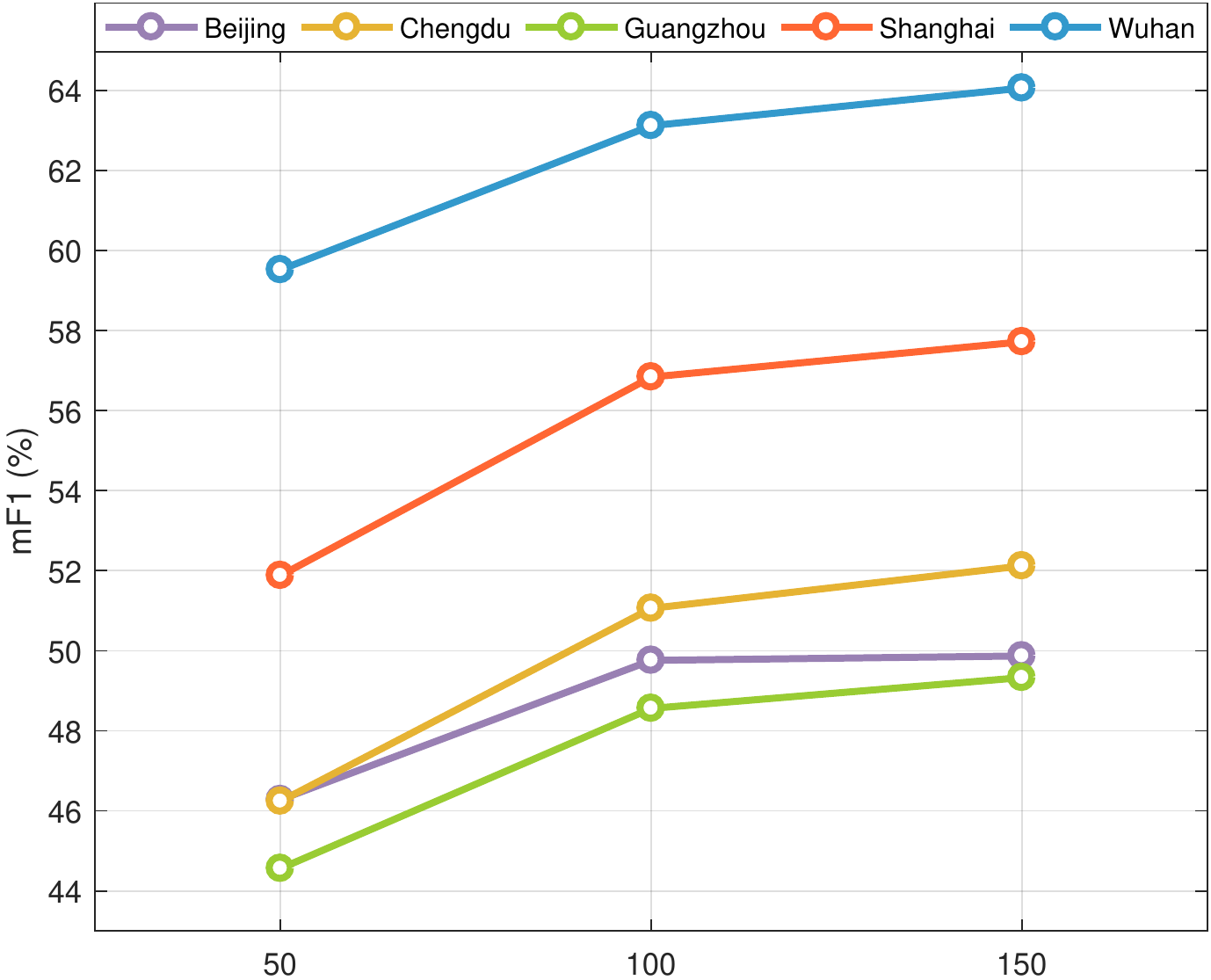}}
\subfigure[\large{$\lambda$}]
{\includegraphics[width=0.3\textwidth]
{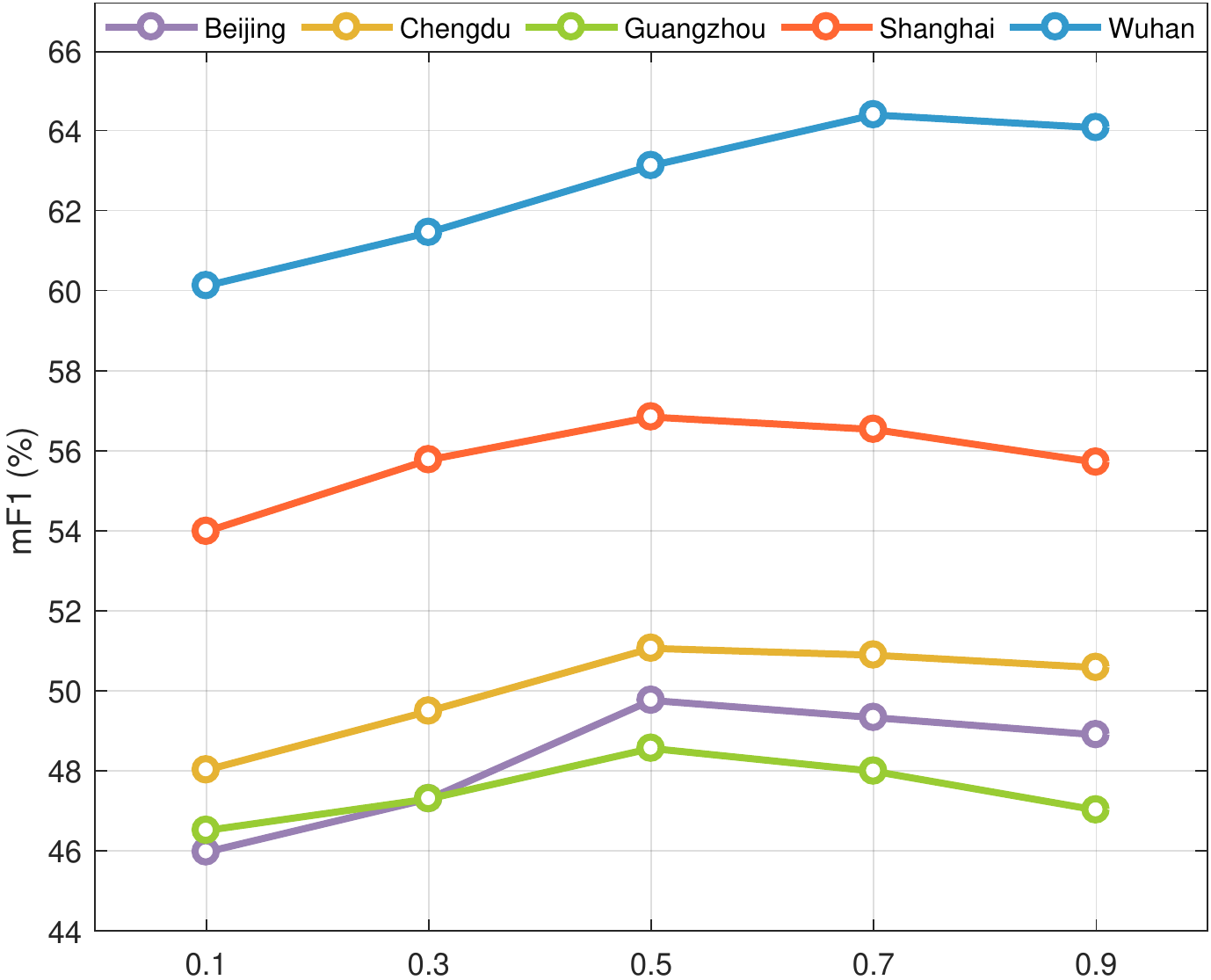}}
\caption{Sensitivity analysis for epoch number and parameter $\lambda$ on Chinese megacities.}
\label{figure:sensitivity}
\end{figure*}

We test three different epoch numbers to study its influence, which are 50, 100, and 150, and the value of $\lambda$ is fixed to 0.5. The relationship between mF1 and the epoch number is presented in Fig. \ref{figure:sensitivity} (a). It can be seen that there is an obvious performance improvement when the epoch number is raised from 50 to 100. However, from 100 to 150, the change in performance flattens out. There are two reasons for this phenomenon: On the one hand, a larger epoch number means fewer pixels are selected at the beginning, which better ensures the quality of the pseudo-labels. On the other hand, a larger number of training iterations enables the model to fully adapt to the target domain. Since 150 epochs would take more computation resources, and the precision improvements are not that significant, 100 epochs are more appropriate in practice.

To investigate how $\lambda$ affects our approach, we test a set of continuously varying values for it, and the epoch number is set to 100. mF1 obtained by each $\lambda$ value is shown in Fig. \ref{figure:sensitivity} (b); it can be seen that the accuracy of each city first rises as $\lambda$ increases and then falls as it becomes larger. The highest accuracy of Wuhan is reached when $\lambda$ value is 0.7, while in other cities it is 0.5. This may be because the sensors of GF-1 and GF-2 have more similar imaging processes, resulting in closer distributions of the source and target domains, and the quality of the pseudo-label can be guaranteed to some extent even if $70\%$ pixels are selected at the end of training, whereas using more pseudo-labels in other cities would introduce more errors. And when $\lambda$ is set to 0.9, $90\%$ pixels in the target domain are involved in the last training epoch, too many errors in pseudo-labels, thus, lead to an obvious decrease in the accuracy of each city.

\section{Discussion}
\subsection{What information is important for land cover classification?}
The built environment and the quality of people's lives are under the combined effects of various land categories \cite{urban}, so it is significant to analyze the land cover information in a more complete category system. However, the recognition of heterogeneous ground objects in high-resolution satellite images is quite difficult. In addition, it depends on different information to identify different categories. For example, classifying independent urban functional buildings relies more on structure and shape features, distinguishing dense built-up areas requires spatial relationships, and the identification of agricultural and natural categories cannot be done without texture and spectral information.

\begin{figure*}[htb!]
\centering
\includegraphics[width=0.7\textwidth]
{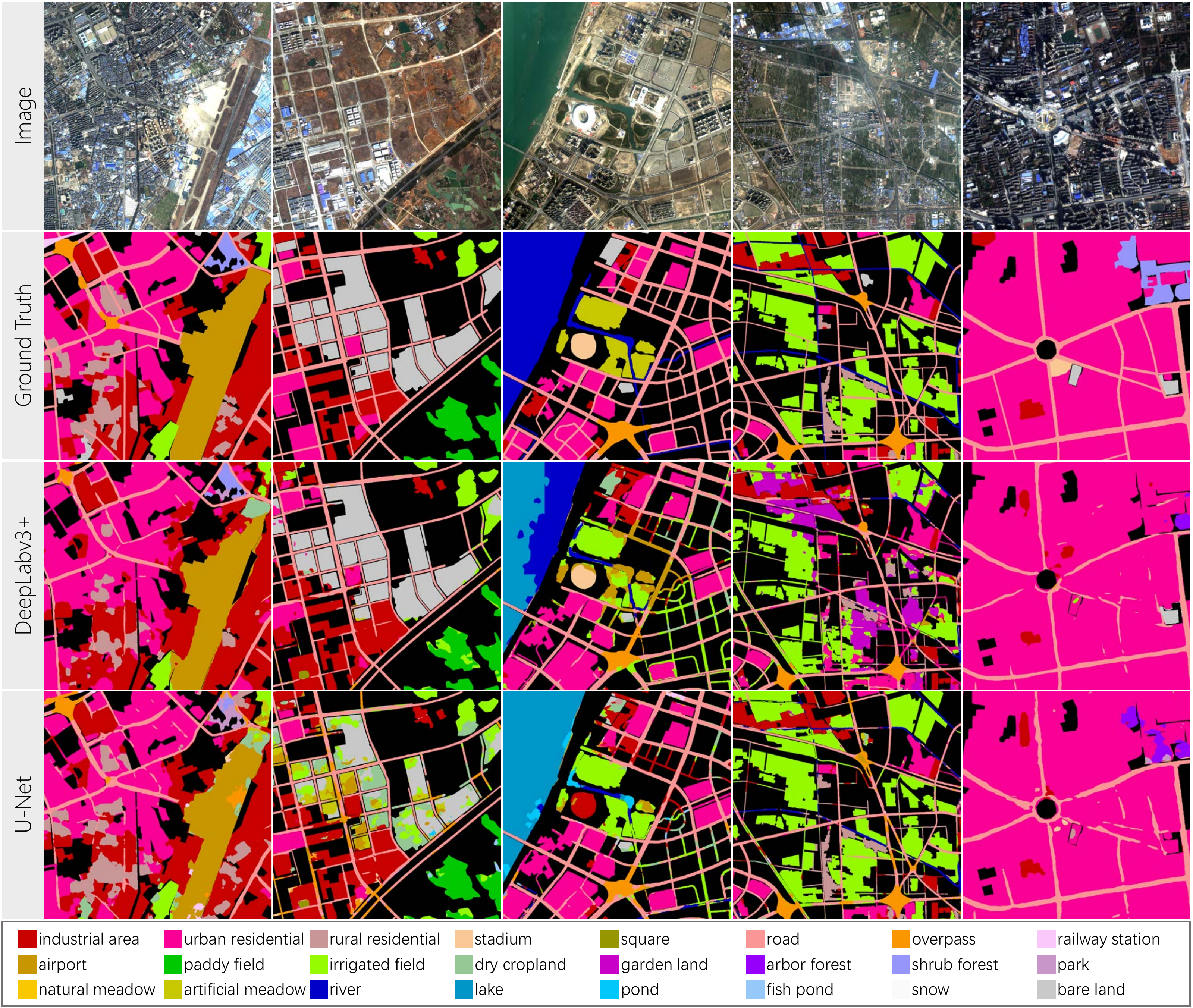}
\caption{Detailed classification results of DeepLabv3+ and U-Net on \emph{Five-Billion-Pixels}. From top to bottom of each column: input satellite image, ground truth, classification result of DeepLabv3+, and classification result of U-Net.}
\label{figure:resultdetail}
\end{figure*}

There is a noteworthy phenomenon in Table \ref{table:resultgid24} and Fig. \ref{figure:resultgid24}. The highest accuracy values for different categories are achieved by different models, mainly DeepLabv3+ and U-Net. We show detailed classification results for \emph{Five-Billion-Pixels} obtained by these two models in Fig. \ref{figure:resultdetail}. In columns 1 to 3, DeepLabv3+ can segment more complete and smooth \emph{airport}, \emph{bare land}, and \emph{stadium}, respectively. And the results of U-Net contain a lot of noise. In visual, DeepLabv3+ tends to identify ground objects as independent ``instances'', whereas it seems difficult for U-Net to aggregate land information into homogeneous segmented regions. This is because the deep features learned by DeepLabv3+ are better at describing contextual and spatial relationships. However, U-Net shows superior performance in columns 4 to 5, where it correctly classifies \emph{irrigated field}, which are partially misclassified by DeepLabv3+ as \emph{garden land} and \emph{road}, which cannot be clearly extracted by DeepLabv3+ from dense built-up areas. This shows that U-Net can more accurately represent the spectra, textures, and sharp boundaries of ground objects.

One explanation for these results is that the ``low-level features'' (those closest to the input image) used in DeepLabv3+’s decoder path are the feature maps that have been forward-propagated through 101 layers and are 16 times smaller compared to the input image. In contrast, U-Net uses ``low-level features'' that go through two layers and have the same scale with the input image since it adopts the concatenation of the encoder and decoder paths. Therefore, U-Net can maintain more raw spectral and edge information to generate dense classification maps.

The above analysis gives us some inspiration, in future research, the combination of ``low-level'' spectral, textural information and ``high-level'' spatial contextual information is likely to facilitate land cover classification in complex category systems.

\begin{figure*}[htb!]
\centering
\subfigure[$\textbf{\emph{D}}_{\textbf{\emph{S}}}$]
{\includegraphics[width=0.2\textwidth]
{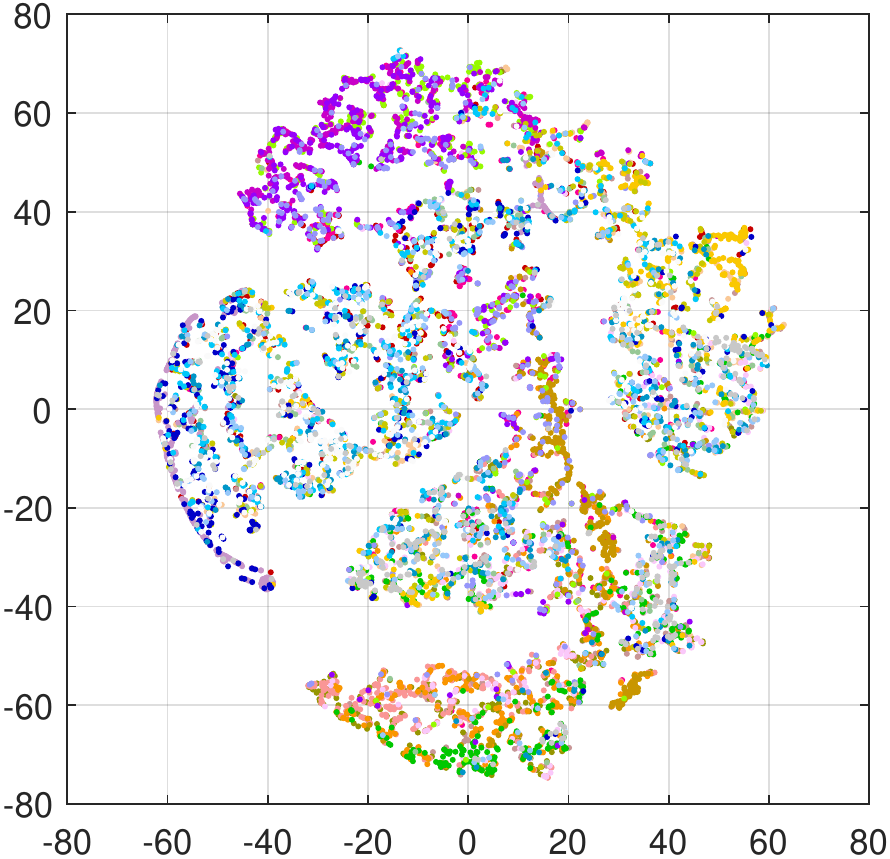}}
\subfigure[Beijing]
{\includegraphics[width=0.2\textwidth]
{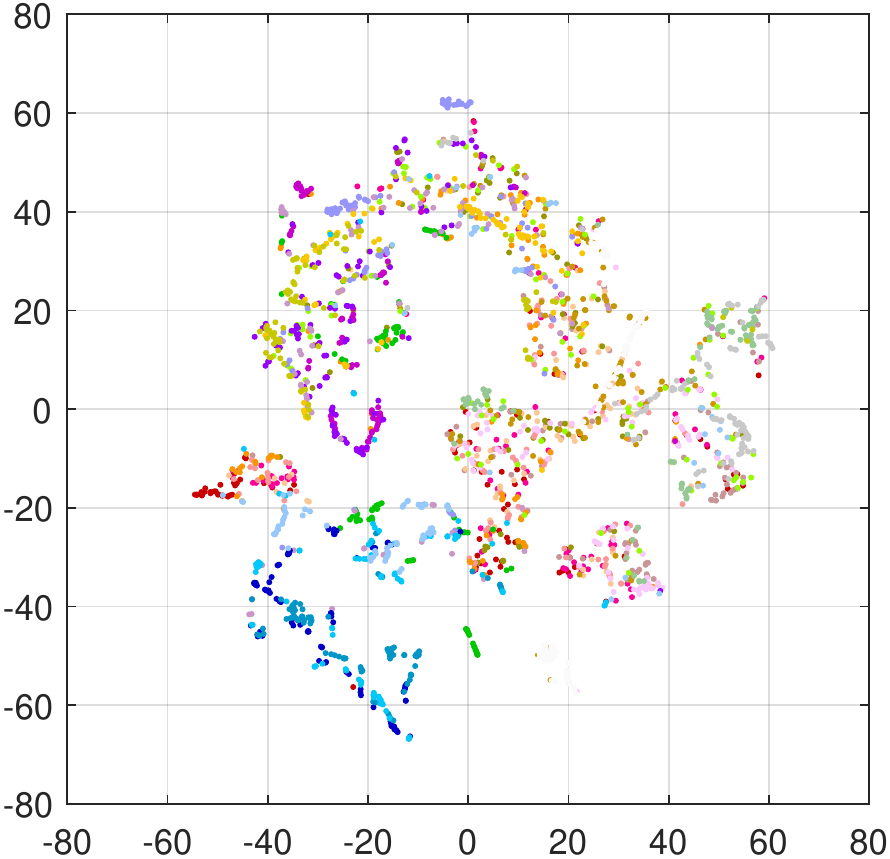}}
\subfigure[Chengdu]
{\includegraphics[width=0.2\textwidth]
{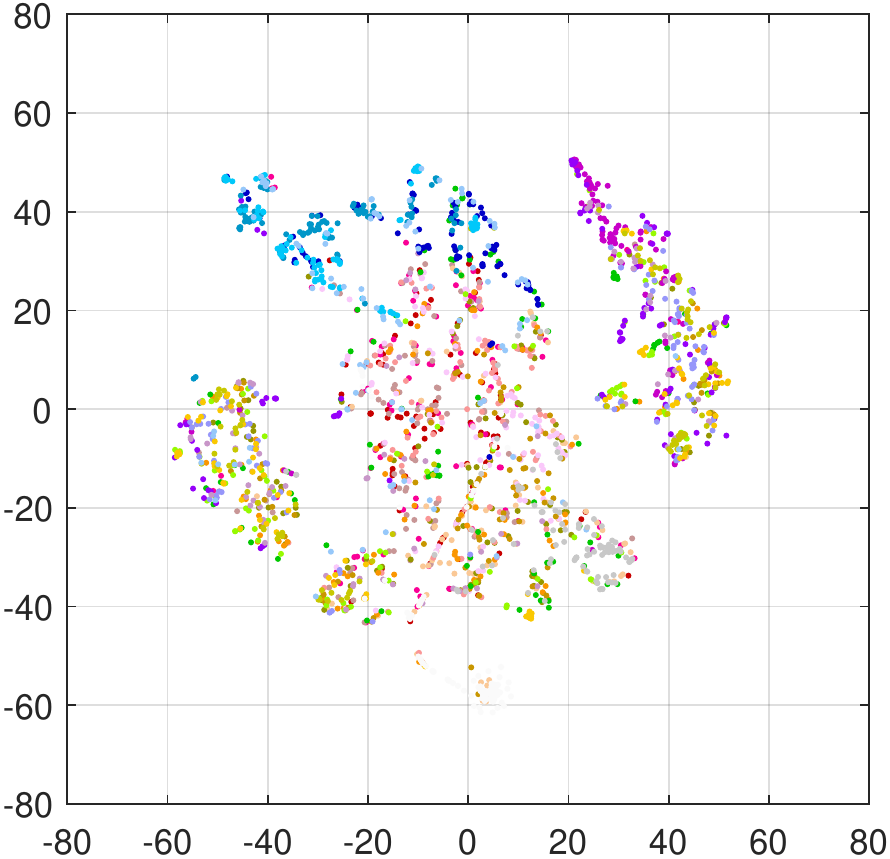}}
\subfigure[Guangzhou]
{\includegraphics[width=0.2\textwidth]
{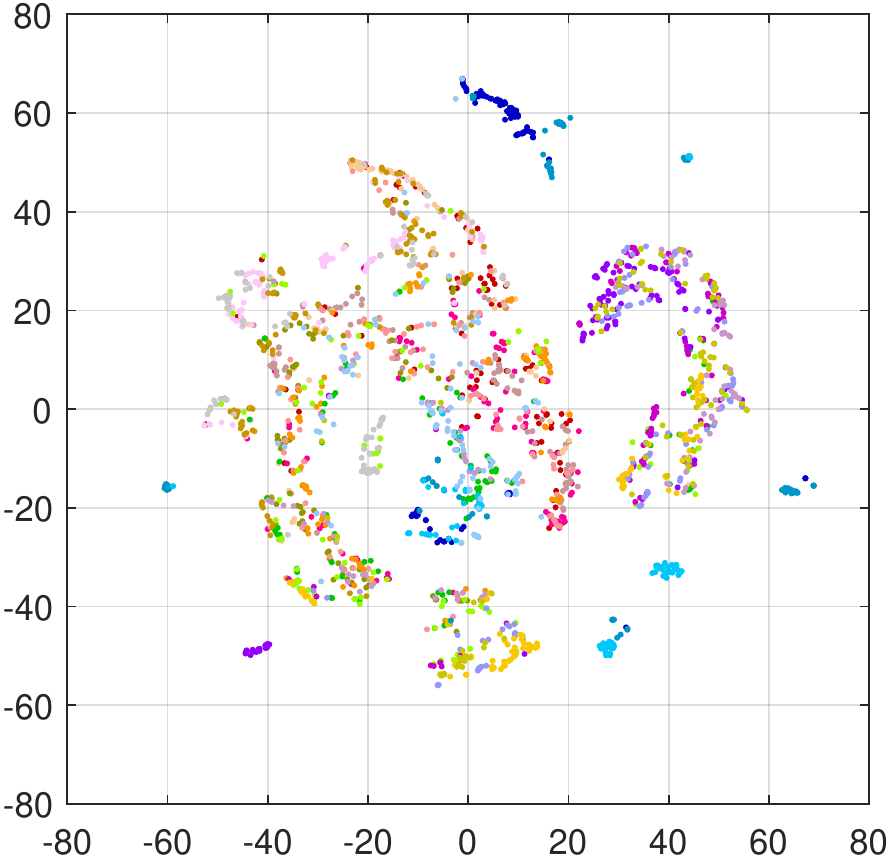}}
\subfigure[Shanghai]
{\includegraphics[width=0.2\textwidth]
{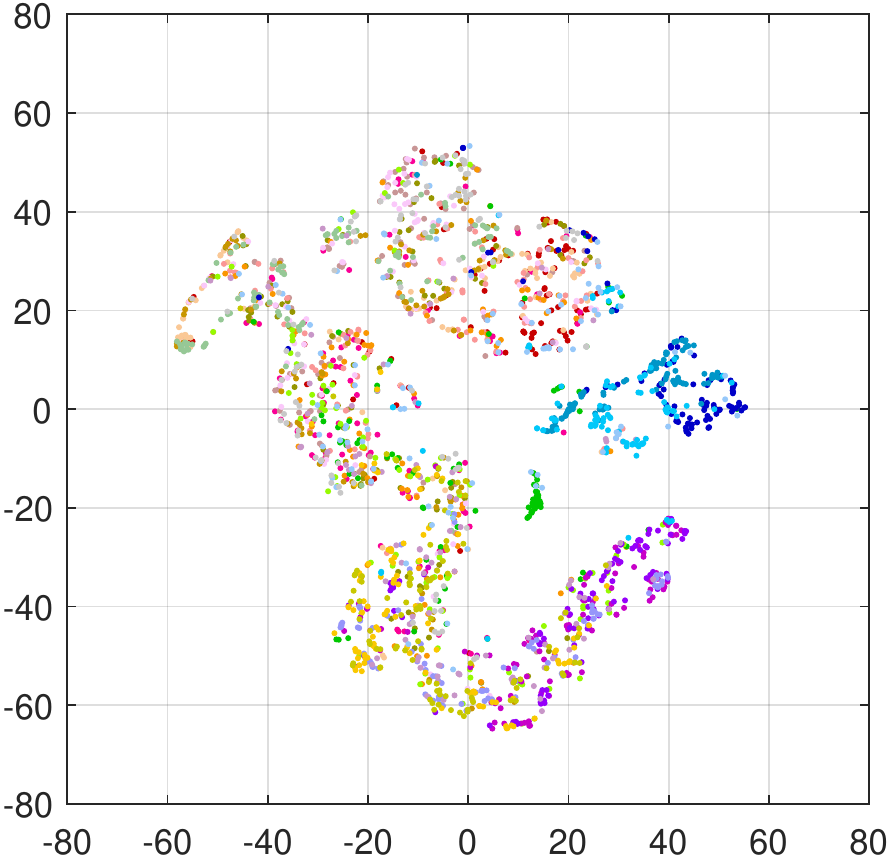}}
\subfigure[Wuhan]
{\includegraphics[width=0.2\textwidth]
{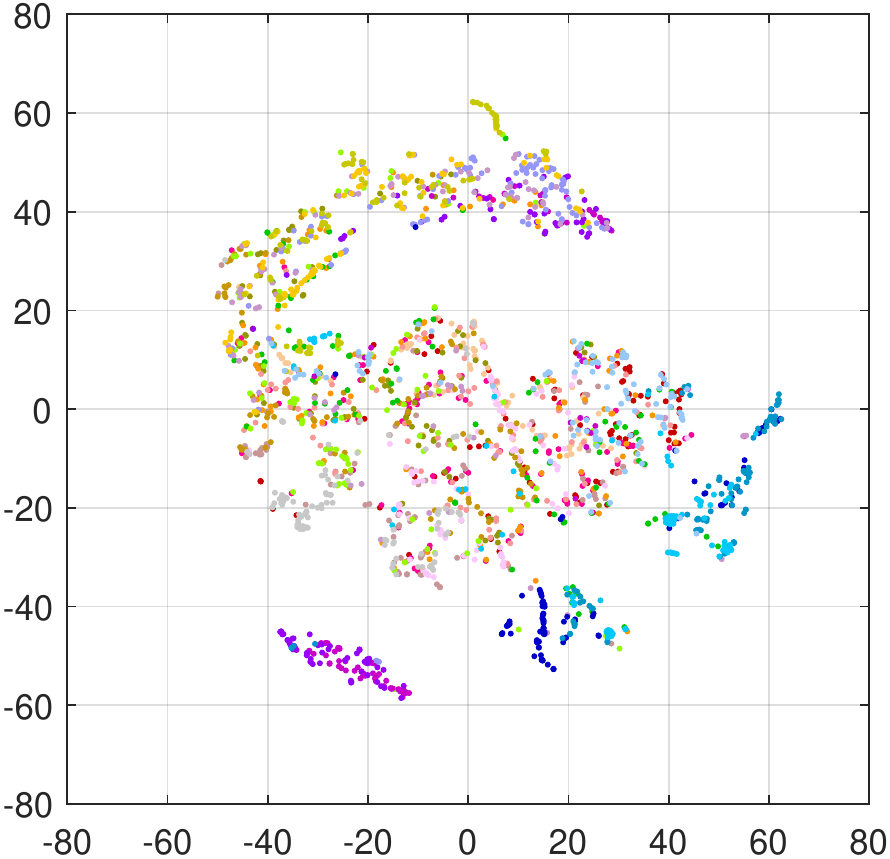}}
\subfigure[Bangkok]
{\includegraphics[width=0.2\textwidth]
{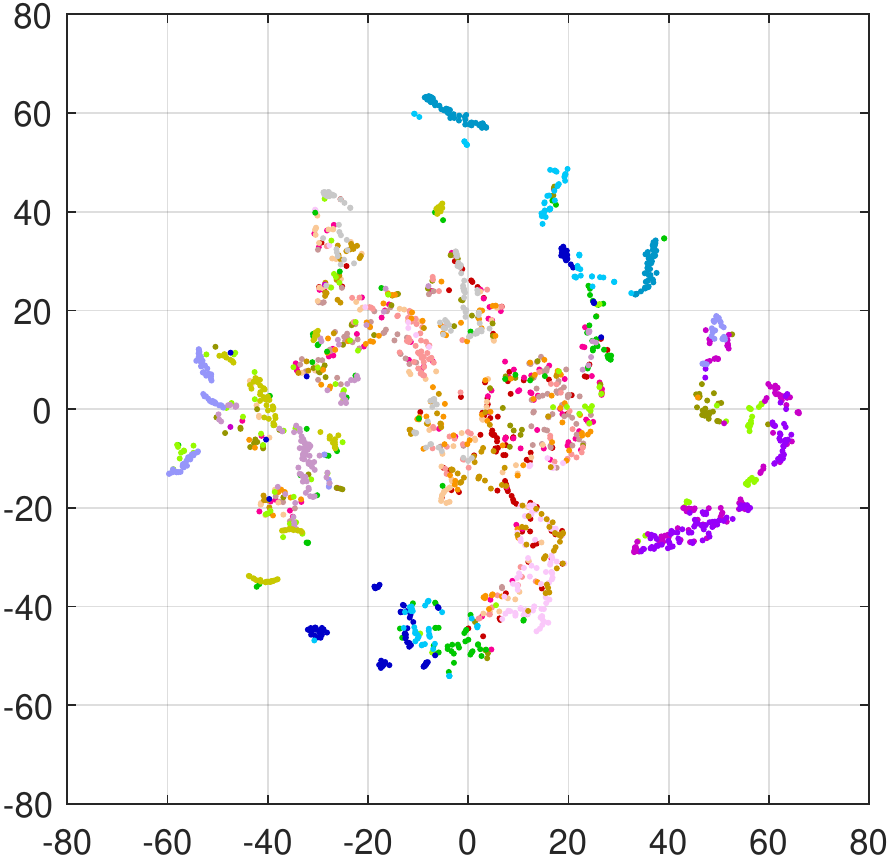}}
\subfigure[Delhi]
{\includegraphics[width=0.2\textwidth]
{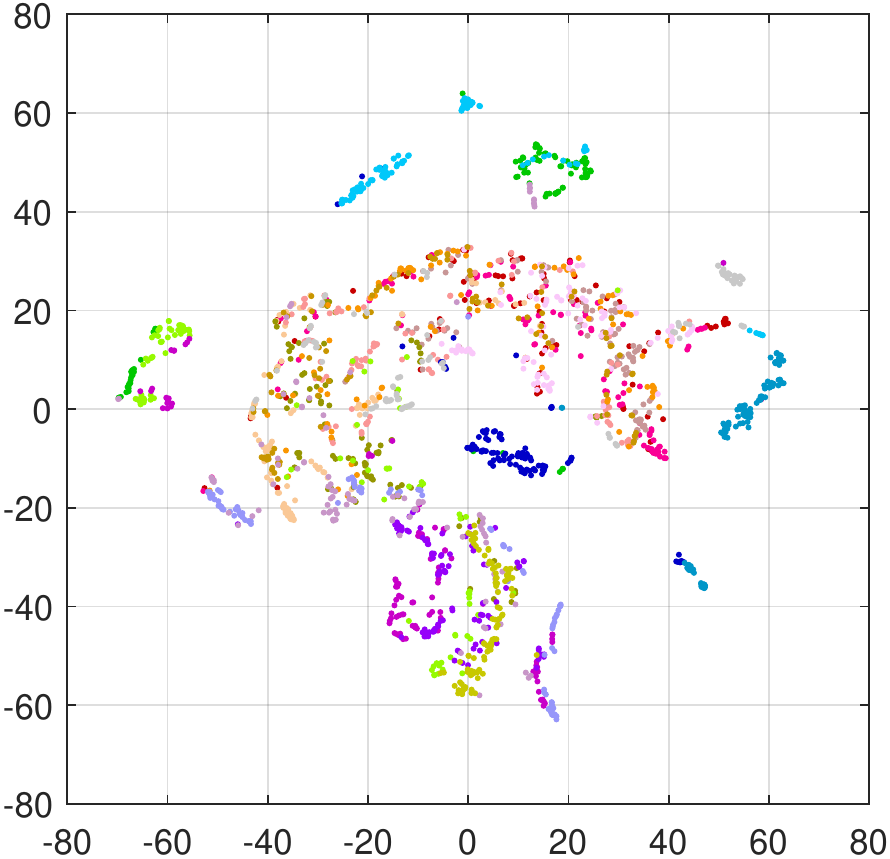}}
\subfigure[Naypyidaw]
{\includegraphics[width=0.2\textwidth]
{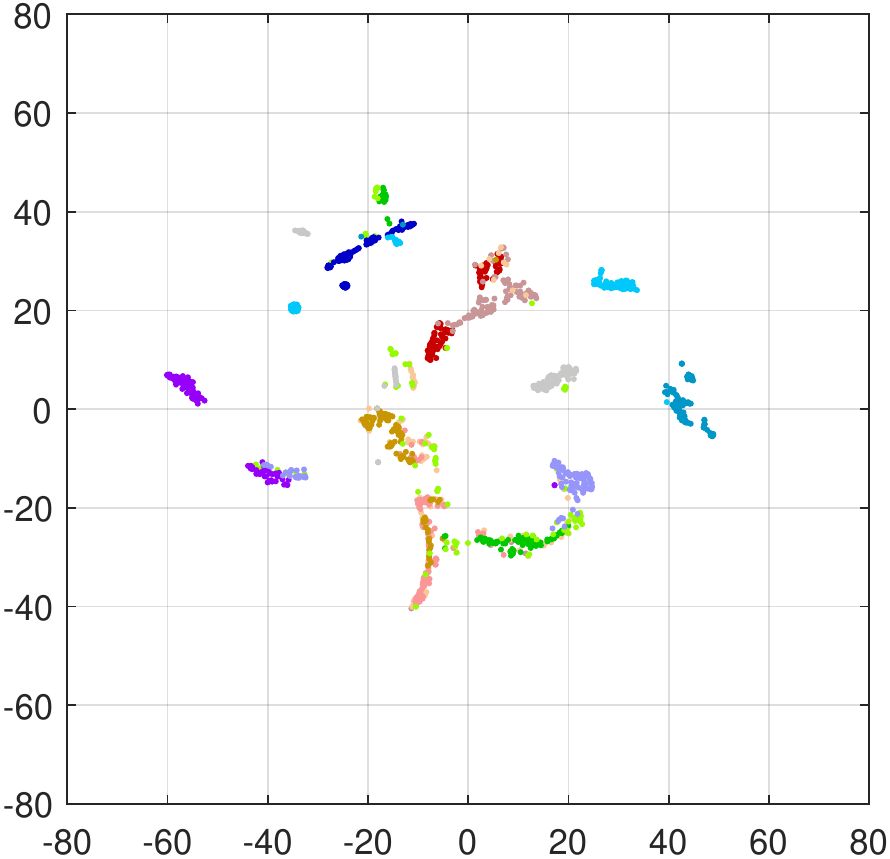}}
\subfigure[Seoul]
{\includegraphics[width=0.2\textwidth]
{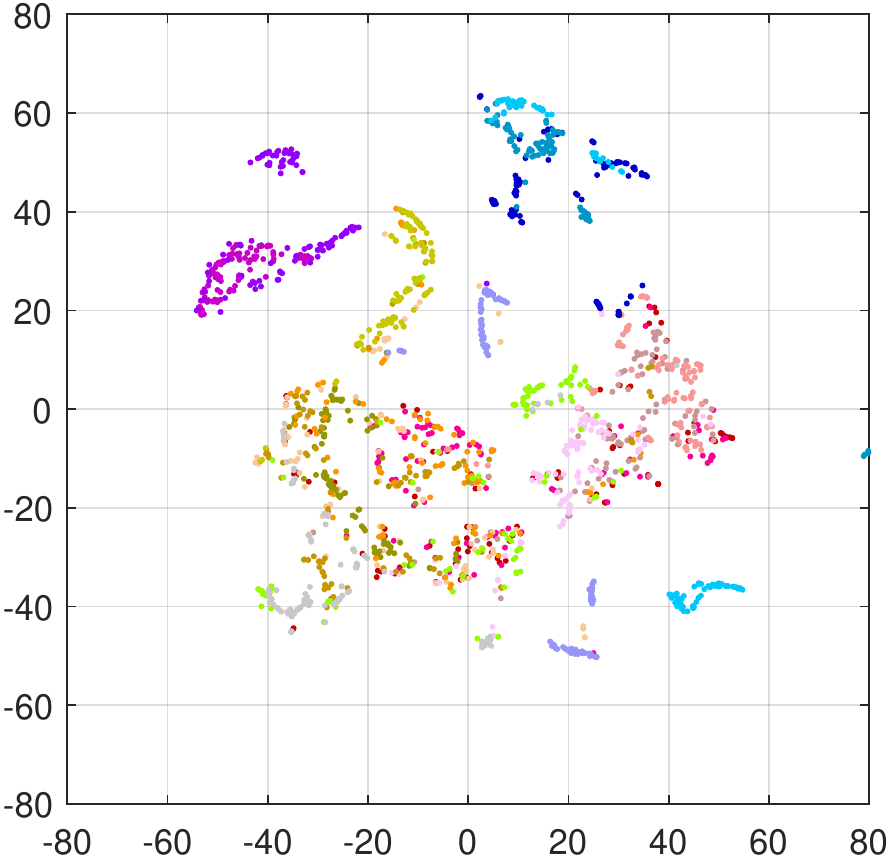}}
\subfigure[Tokyo]
{\includegraphics[width=0.2\textwidth]
{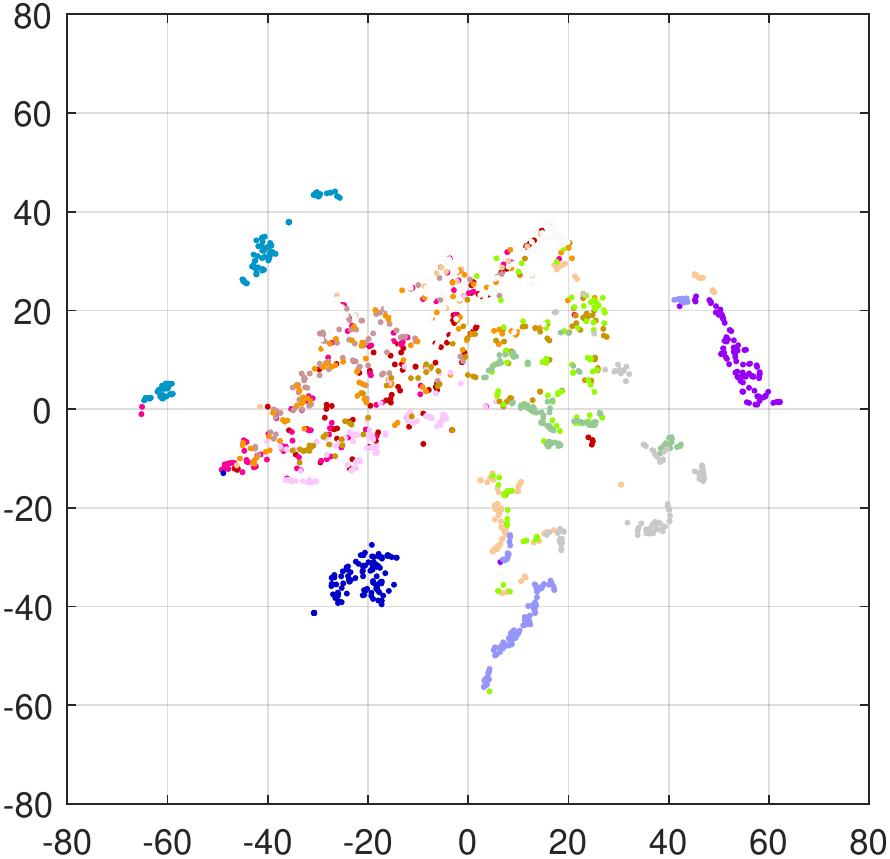}}
\subfigure[Yangon]
{\includegraphics[width=0.2\textwidth]
{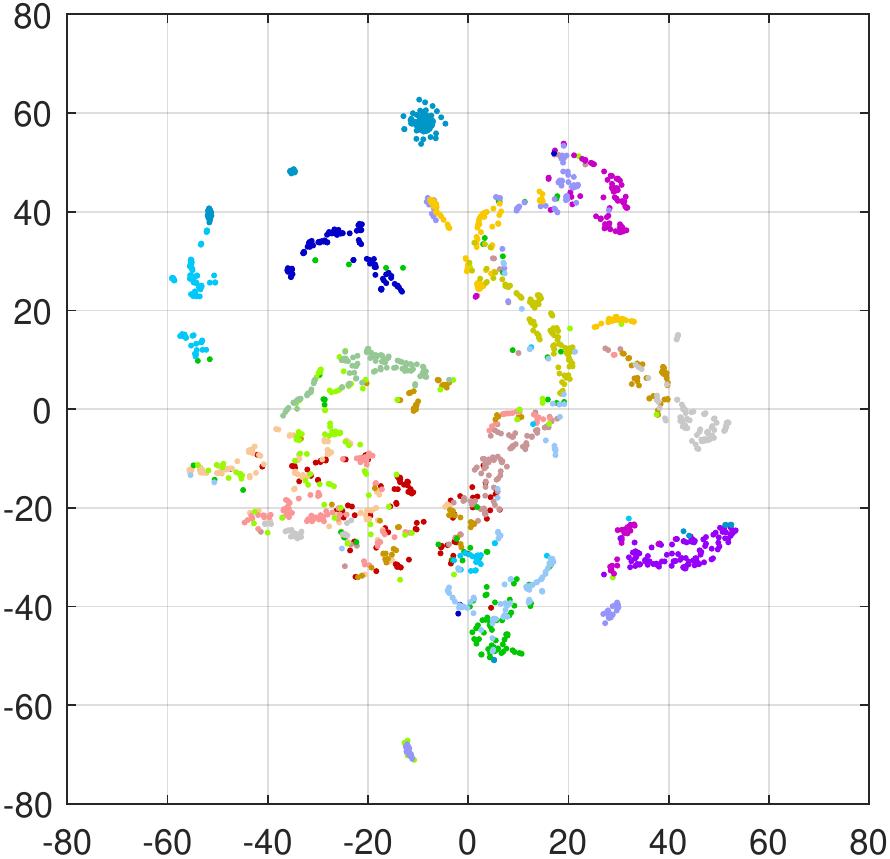}}
\caption{Feature spaces of different domains (\emph{Five-Billion-Pixels} and 11 Asian cities). For $\textbf{\emph{D}}_{\textbf{\emph{S}}}$, 400 samples per category are presented. For each $\textbf{\emph{D}}_{\textbf{\emph{T}}}$, 100 samples per category are displayed.}
\label{figure:featurespace}
\end{figure*}

\subsection{How can the performance of land cover mapping be further improved?}
As can be seen in Fig. \ref{figure:resultlarge}, while our approach performs well in most areas of each Chinese megacity, the results for mountainous forest surrounding the cities, especially Beijing and Chengdu, are not that satisfactory. To discuss this phenomenon, we visualize the feature spaces of different data domains with t-SNE \cite{t-SNE} in Fig. \ref{figure:featurespace}, where the combination of spectral features and texture features (GLCM) is employed, and the coordinate systems of the feature spaces are aligned.

Three characteristics of feature distributions can be observed from Fig. \ref{figure:featurespace}: (1) distribution shifts occur between different domains; (2) within each domain, the distribution of almost every category is dispersive; (3) within each domain, the distribution of different categories may be partially mixed. The last case is evident in forest and cropland categories of $\textbf{\emph{D}}_{\textbf{\emph{S}}}$. This is because their features are largely influenced by geographical location and seasonal changes. For example, the \emph{Five-Billion-Pixels} dataset covers a large amount of \emph{irrigated field} reclaimed in the mountains of northwestern China, and their spectral and texture features may be similar to those of mountainous \emph{arbor forest} in winter. Then, in the process of domain joint learning, the DCNNs model is likely to assign \emph{irrigated field} pseudo-labels to a small number of \emph{arbor forest} samples in $\textbf{\emph{D}}_{\textbf{\emph{T}}}$. And these errors will further accumulate in continuous iterative training, eventually leading to misclassification in the land cover mapping results.

Since this problem is caused by the inherent constraints of UDA and the intrinsic properties of vegetation categories, in future studies, the integration of other sources of information may lead to improvement. For instance, the all-season sample dataset \cite{2017first} provides vegetation samples from multiple seasons at 30 m resolution, and multi-temporal analysis methods \cite{2016evaluation,2018novel} can better distinguish the coverage of different vegetation categories. It is promising to obtain more accurate, high-resolution mapping results for both natural and urban areas by multi-source data fusion \cite{2018mapping,mediumETM+} using the \emph{Five-Billion-Pixels} dataset.

\subsection{What can be further done based on our dataset and approach?}
We achieve fully automated classification with unlabeled high-resolution satellite images, which opens up new possibilities for large-scale, real-time land cover mapping. Moreover, our approach is proven to be generalizable across images captured by different sensors, especially for ST-2 images that are free and open. As a continuous, reliable, quality-controlled data source, Sentinel satellite data are utilized in the state-of-the-art global land cover mapping projects, ESA’s World Cover \cite{esaLC} and Google's Dynamic World \cite{googleLC}. Since there is some comparability at the same spatial resolution, we compare our results of Beijing and Guangzhou with Dynamic World and World Cover.

\begin{figure*}[htb!]
\centering
\includegraphics[width=0.96\textwidth]
{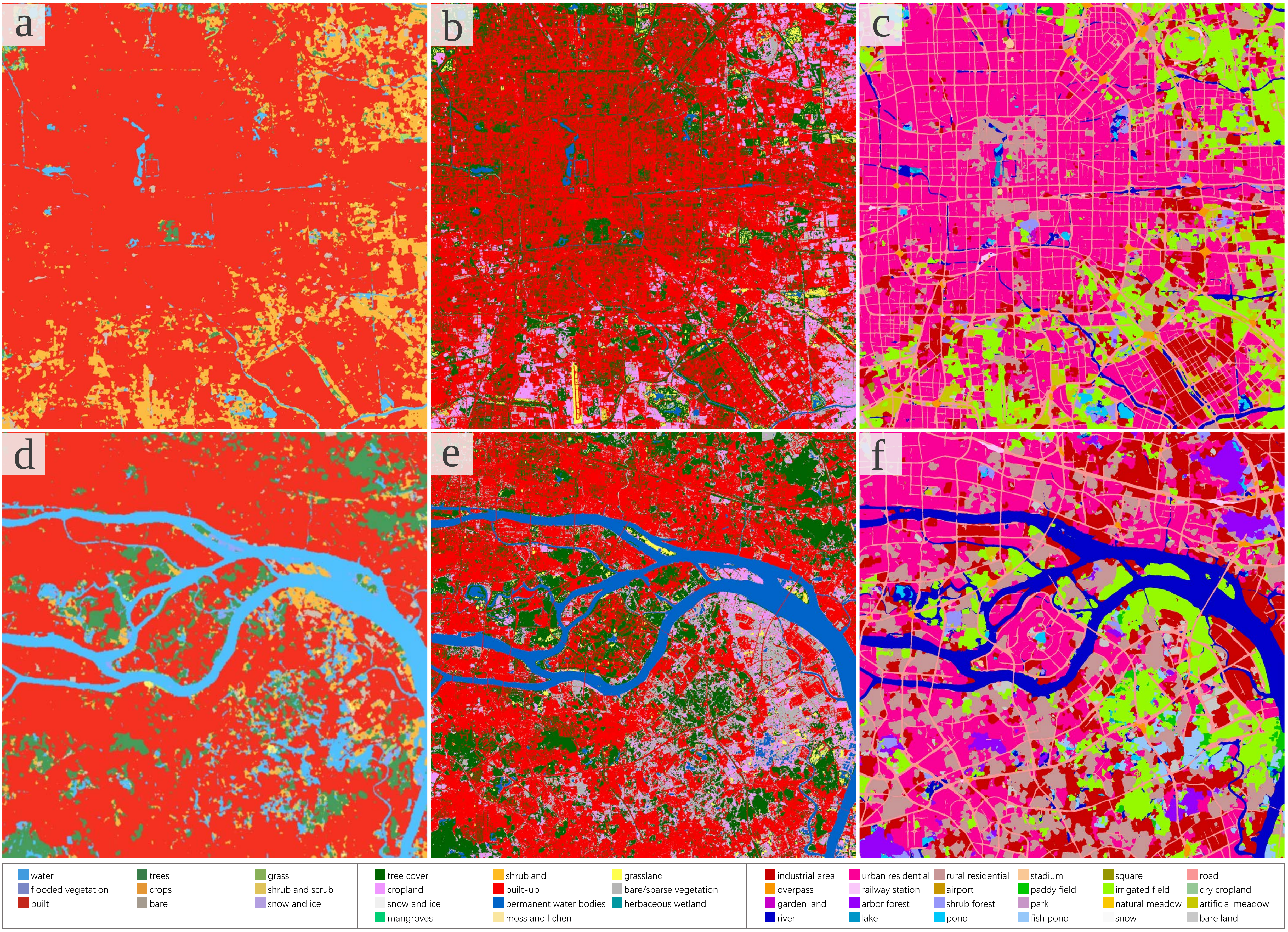}
\caption{Comparison with Google's Dynamic World and ESA's World Cover in Beijing (ST-2) and Guangzhou (ST-2). (a-b) Land cover map of central area of Beijing from Dynamic World and World Cover. (c) Our result of Beijing. (d-e) Land cover map of central area of Guangzhou from Dynamic World and World Cover. (f) Our result of Guangzhou.}
\label{figure:resultcmpr}
\end{figure*}

Due to the different acquisition time of the original images, we can only make an approximate visual comparison. As shown in Fig. \ref{figure:resultcmpr}, our results are able to distinguish forest, built-up, water bodies, and cropland that are relatively consistent with Dynamic World and World Cover. Furthermore, owing to our extensive, fine, and accurate annotating for \emph{Five-Billion-Pixels}, our results present clear transportation networks and river systems, as well as different agricultural and urban functional areas. It can be seen that, because of the rapid urban development in China, new high-rise residential buildings are mixed with old ones, which have the same appearance as rural settlements, creating a special urban landscape. The richer categories of our results have the potential to contribute to studies on urban planning, urban heat islands, urban quality of life, and so on.

Although we studied only 11 cities in this paper, our approach can be easily generalized to other cities, towns, and villages throughout China and even other Asian countries. In addition, the discrete pseudo-labels used in our UDA approach are validated to be capable of improving the classification results. This suggests that the annotation of newly acquired images do not need to be dense and pixel-wise when a large-scale, well-annotated dataset is already available. Therefore, for other countries and regions with very different land distributions and land category systems than China, it has potential to perform land cover mapping based on \emph{Five-Billion-Pixels} and, for example, semi-supervised domain adaptation with sparse annotations in the form of patches or polygons. This is an issue of interest to us in the future.

\section{Conclusion}
The increasing volume of high-resolution satellite data is a ``gold mine’’ waiting to be explored and mined. Yet land cover mapping on a large-scale in high-resolution remains a challenging task. In this paper, we present a large-scale land cover dataset, \emph{Five-Billion-Pixels}, which can provide the remote sensing community with a high-quality benchmark to advance land cover classification algorithms. At the same time, we propose an unsupervised domain adaptation approach that can deal with complicated real-world distribution shifts. The land cover mapping results for five megacities in China and six cities in other Asian countries show the generalizability of our approach across different sensors and geographical regions. In general, our work has the potential to be extended to land cover mapping at country-scale and to contribute to various applications involving land cover information.

\bibliographystyle{IEEEtran}
\bibliography{reference}

\end{document}